%% file: main.tex
\documentclass{article}
\usepackage{recon,times}

\usepackage[colorlinks=true,linkcolor=red,citecolor=green!50!black,urlcolor=blue]{hyperref}
\newcommand{\figpart}[2]{\hyperref[#1]{\ref*{#1}#2}}
\usepackage{url}
\usepackage{graphicx}
\usepackage{subcaption}
\usepackage{amsmath,amssymb}
\usepackage{booktabs}
\usepackage{multirow}
\usepackage[table]{xcolor}
\usepackage{tikz}
\usepackage{needspace}
\usepackage{wrapfig}
\usepackage{enumitem}
\usepackage{xcolor}
\usepackage{soul}
\sethlcolor{black!10}
\usepackage{algorithm}
\usepackage{listings}

\definecolor{codecomment}{rgb}{0.25,0.5,0.5}
\definecolor{codefunc}{rgb}{0.85,0.18,0.50}
\definecolor{codeop}{rgb}{0,0,1}
\definecolor{cmethod}{gray}{0.4}
\lstdefinestyle{pseudo}{
  basicstyle=\fontsize{8pt}{9.6pt}\ttfamily\selectfont,
  columns=fullflexible,
  keepspaces=true,
  mathescape=true,
  morecomment=[l]{\#},
  commentstyle=\color{codecomment},
  emphstyle=\color{codefunc},
  aboveskip=4pt,
  belowskip=0pt,
  literate=*{+}{{\textcolor{codeop}{+}}}1
            {-}{{\textcolor{codeop}{$-$}}}1
            {*}{{\textcolor{codeop}{$\ast$}}}1
            {/}{{\textcolor{codeop}{/}}}1,
}
\definecolor{cdepth}{RGB}{235,235,255}
\definecolor{ctok}{RGB}{229,255,229}
\definecolor{ctmax}{RGB}{255,254,230}
\definecolor{cdm}{RGB}{255,235,235}
\definecolor{cscale}{RGB}{255,254,230}
\definecolor{ctokz}{RGB}{235,235,255}
\definecolor{cenc}{RGB}{229,255,229}
\DeclareRobustCommand{\hlwrap}[2]{{\sethlcolor{#1}\hl{#2}}}

\usetikzlibrary{arrows.meta,calc}

\definecolor{patchtok}{HTML}{A8B2BD}
\definecolor{semtok}{HTML}{F3E3A6}
\definecolor{enctok}{HTML}{AFCBE9}
\definecolor{trainpath}{HTML}{3C8C3C}
\definecolor{evalpath}{HTML}{E03A3A}

\definecolor{concatarrow}{HTML}{2E6C8A}
\DeclareCaptionFont{eightpt}{\fontsize{8}{9.5}\selectfont}

\newcommand{\vary}[2][gray!18]{\cellcolor{#1}#2}%

\usepackage{bm}

\title{Representation Conditioning for Generation}

\author{Saeed Razavi$^{1}$, Xingjian Leng$^{1}$, Liang Zheng$^{1}$ \\
$^{1}$Australian National University \\
\texttt{\{Seyedsaeed.Razavi,Xingjian.Leng,liang.zheng\}@anu.edu.au}}

\begin{document}

\maketitle

\vspace{-12pt}
\begin{abstract}
Representation-alignment methods effectively accelerate the diffusion transformer training. REPA, for example, uses a pretrained
vision encoder to supervise an intermediate layer through an {MLP projection head} which is discarded after training.
It aims to import semantics from the vision encoder to the diffusion backbone. While this objective is achieved at \textit{early layers} by back-propagating the alignment loss, there is no guarantee that \textit{later layers}, supervised by the velocity prediction loss alone, can well access such semantics. We propose a simple solution, representation conditioning (RECON), to improve semantics in later layers. During training, we feed the pretrained vision encoder tokens into later layers; during inference, because we no longer have the clean image, we reuse the MLP head (as a proxy to the vision encoder) and project its output tokens to the later layers. Our method
adds no parameters and leads to much faster convergence: RECON is
$63\times$ faster than SiT and $25\times$ faster than REPA.
Code is available at \url{https://github.com/saeidrazavi/RECON}.

\end{abstract}

\begin{figure}[!ht]
\centering
\captionsetup[subfigure]{font=eightpt,justification=centering,skip=0pt}
\begin{tikzpicture}[inner sep=0]
  \def\panelh{124.5pt}
  \def\gap{9pt}
  \node[anchor=south west] (repa) at (0,0)
    {\includegraphics[page=3,height=\panelh,trim=13 196 762.26 103,clip]%
       {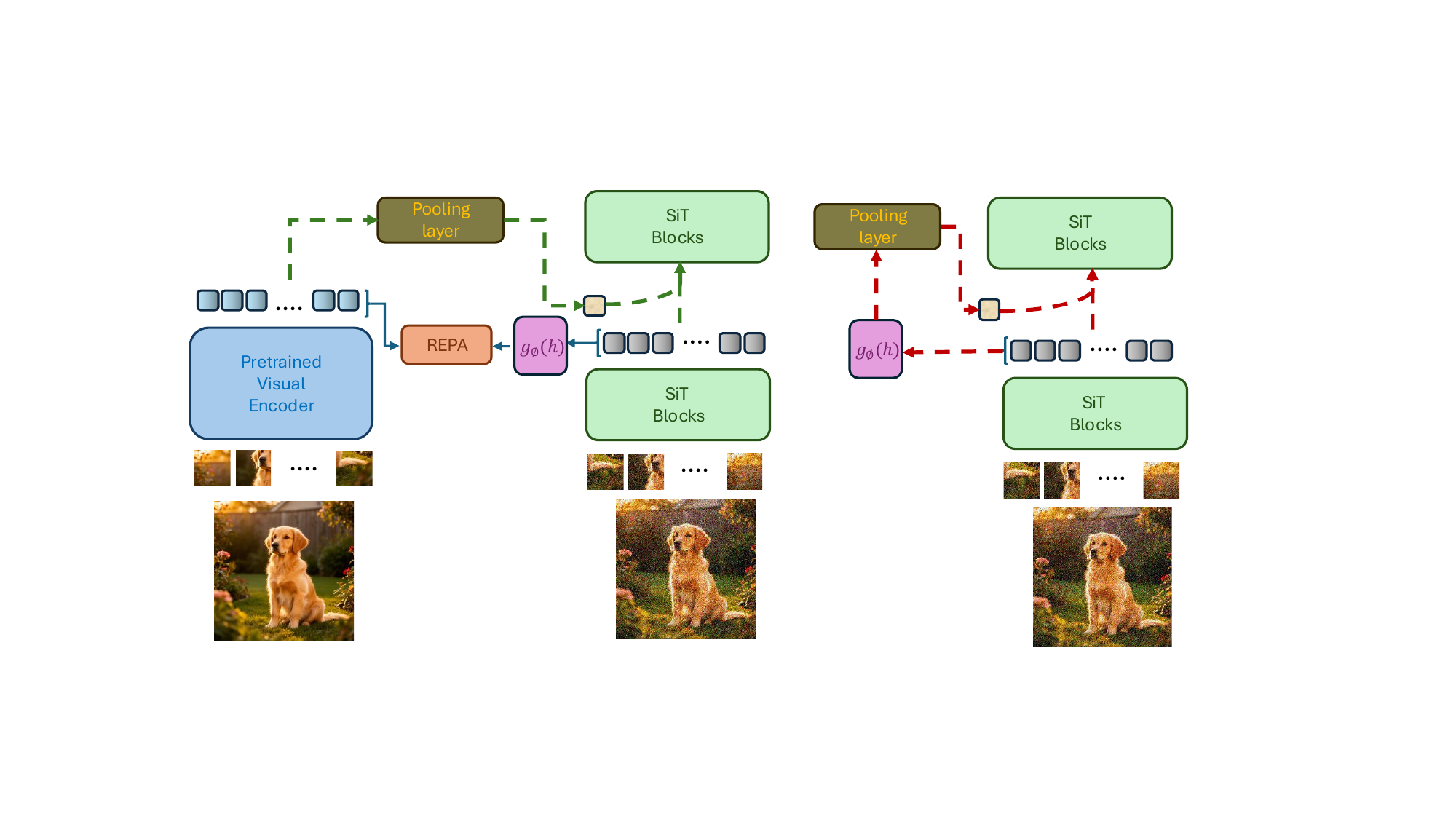}};
  \node[anchor=south west] (train) at ([xshift=\gap]repa.south east)
    {\includegraphics[page=3,height=\panelh,trim=209.5 196 583.76 103,clip]%
       {figures/Presentation2_minipage_v2.pdf}};
  \node[anchor=south west] (infer) at ([xshift=\gap]train.south east)
    {\includegraphics[page=3,height=\panelh,trim=396 196 424.26 103,clip]%
       {figures/Presentation2_minipage_v2.pdf}};
  \node[anchor=south east] (plot) at (\linewidth,0)
    {\includegraphics{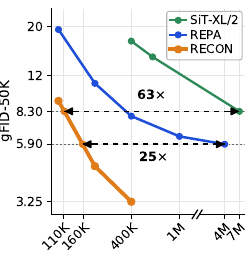}};
  \foreach \pnl/\sub in {repa/REPA, train/RECON training,
                         infer/RECON inference}
    \path let \p1 = ($(\pnl.south east)-(\pnl.south west)$) in
      node[anchor=north,text width=\x1,align=center] (\pnl-cap)
        at ([yshift=-4pt]\pnl.south) {\subcaption{\sub}};
  \path let \p1 = ($(plot.south east)-(plot.south west)$) in
    node[anchor=north,text width=\x1,align=center]
      at ([yshift=-4pt]plot.south)
      {\subcaption[Training steps vs.\ gFID-50K]%
         {Training steps vs.\ gFID-50K\protect\\
          Improved generation performance}};
  \path ([xshift=\gap/2]repa.north east) coordinate (d1)
        ([xshift=\gap/2]train.north east) coordinate (d2);
  \draw[gray!65,line width=0.6pt] (d1) -- (d1 |- train-cap.south);
  \draw[gray!65,line width=0.6pt,dashed] (d2) -- (d2 |- infer-cap.south);
\end{tikzpicture}
\caption{\textbf{RECON conditions the later layers on the representation that
REPA discards.} (a) REPA trains a projection head $g_\phi$ against a
frozen vision encoder through the REPA loss, then discards it at inference. RECON inherits most of REPA, but (b) conditions later layers with the vision encoder tokens during training and (c) during inference retains $g_\phi$ as a proxy to the vision encoder and does the same conditioning.
(d) gFID-50K against training
step on ImageNet $256\times256$ without guidance; RECON reaches the final quality of SiT-XL/2 and of REPA much earlier.}
\label{fig:teaser}
\end{figure}

\begin{figure}[t]
\centering
\captionsetup[subfigure]{font=normalsize,justification=centering,skip=0pt}
\raisebox{-15.19pt}{\includegraphics[scale=0.872,trim=1.782 3.240 1.039 2.224,clip]{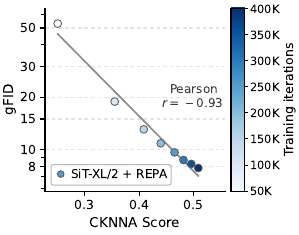}}\hfill
\raisebox{-16.29pt}{\includegraphics[scale=0.872,trim=1.692 1.980 2.708 4.360,clip]{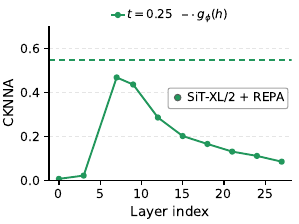}}\hfill
\raisebox{-20.47pt}{\includegraphics[scale=0.872,trim=2.592 2.880 1.976 1.976,clip]{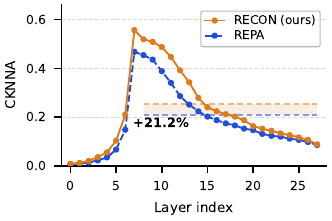}}\par\vspace{3pt}
\begin{subfigure}[t]{122.3pt}\subcaption{Strong correlation of gFID and CKNNA}\label{fig:corr}\end{subfigure}\hfill
\begin{subfigure}[t]{120.9pt}\subcaption{$g_\phi$ output has more semantics than any layer}\label{fig:cknaa}\end{subfigure}\hfill
\begin{subfigure}[t]{134.7pt}\subcaption{RECON improves CKNNA across layers}\label{fig:dynamics_cknna}\end{subfigure}
\caption{\textbf{Alignment with the vision encoder, measured by CKNNA
against DINOv2-B.} (a) gFID against CKNNA over training checkpoints. (b) CKNNA of backbone features at each layer. Dashed line is CKNNA between the projector output $g_\phi(h)$ and the vision encoder, which is stronger than any layer. (c) CKNNA of every layer for REPA and RECON at $t=0.25$.}
\label{fig:alignment}
\end{figure}

\section{Introduction}
\label{sec:intro}

An effective strategy for efficient diffusion transformer training is to align the diffusion features with an external vision encoder \citep{yu2025repa}.
Specifically, REPA \citep{yu2025repa}
uses a {lightweight projection head $g_\phi$}\footnote{We use $g_\phi$ interchangeably with ``projection head''.%
\hfill\textbf{Project page:} \href{https://recondiffusion.github.io}{\texttt{recondiffusion.github.io}}} to map the intermediate features of diffusion backbone to be similar to those of the frozen vision encoder (Fig.~\figpart{fig:teaser}{a}).
Several follow-up works build on this idea, \textit{e.g.}, REG \citep{wu2025reg},
DDT \citep{wang2025ddt} and REPA-E \citep{leng2025repae}.

By aligning the diffusion model with an external vision encoder, the assumption is that the diffusion backbone will \textit{effectively} access and use the semantic features, so that generation will eventually improve. In this work, we examine the alignment between the vision encoder features and features of each layer of the diffusion transformer, and the feature from $g_\phi$. We find that the feature from $g_\phi$ has the highest level of alignment with the vision encoder, as shown in Fig.~\figpart{fig:alignment}{b}. This finding is not surprising, because $g_\phi$ is physically closest to the vision encoder and is directly optimized to minimize alignment loss.

Surprisingly, despite its strong alignment with the vision encoder, existing methods typically discard the projection head $g_\phi$ after training, as shown in Fig.~\figpart{fig:teaser}{a}. We argue that the practice is a waste of model resources including learned weights and semantics: the full potential of representation alignment is not achieved.

We introduce a very simple fix, named REpresentation CONditioning (RECON). Our idea is that instead of removing $g_\phi$ after training, we keep it and use it as a proxy of the vision encoder during inference, so that strong semantics from $g_\phi$ can be explicitly used to benefit generation.  Specifically, during inference, RECON pools the $g_\phi$ output features into \textit{semantic tokens}, concatenates
them with the patch tokens, and lets the diffusion backbone attend to both types of tokens. Correspondingly, we have a small modification for training recipe: during training, the semantic tokens directly come from the vision encoder and serve as condition for the backbone.

Despite its simplicity, this fix effectively improves the semantics of the backbone features, as shown in Fig.~\figpart{fig:alignment}{c}. Meanwhile, this method also improves generation quality: we report FID = 3.25 at 80 epochs using SD-VAE, which is competitive with the state-of-the-art methods. This is consistent with previous observation that better semantic alignment leads to better generation quality (Fig.~\figpart{fig:alignment}{a}). Furthermore, our method is advantageous in that it adds no parameters\footnote{It generally holds for B, L, and XL models. For certain S models, we add a linear layer. See Sec.~\ref{sec:training}.} and needs fewer sampling steps to reach the same quality (Table~\ref{tab:steps}). We summarize the main points of this paper below.

\begin{figure}[t]
\centering
\includegraphics[width=\linewidth]{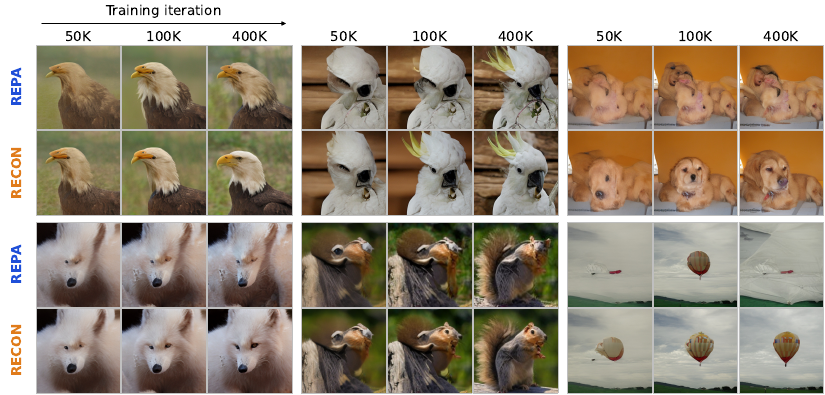}
\caption{\textbf{Representation conditioning improves visual scaling.} RECON produces higher-quality images at different
steps compared with vanilla REPA. Results for
both methods are sampled using the same seed, noise and class label. We use no classifier-free guidance.}
\label{fig:quality}
\end{figure}

\begin{itemize}[leftmargin=*,itemsep=1pt,topsep=2pt]
\item We question the common practice of discarding $g_\phi$ after training. Retaining it allows the semantics learned through representation alignment to guide generation.
\item We introduce RECON that reuses $g_\phi$ at inference
to make the semantics more accessible (Fig.~\figpart{fig:teaser}{c}).
\item RECON converges $63\times$ faster than SiT and $25\times$ faster than
REPA (Fig.~\figpart{fig:teaser}{d}, Fig.~\ref{fig:quality}) and is parameter efficient and sampling-step efficient.
\end{itemize}

\section{Related Work}
\label{sec:related}

\textbf{Representation alignment.} REPA \citep{yu2025repa} is a representative work.
DDT \citep{wang2025ddt} builds the separation into the architecture,
dividing the transformer into a semantic encoder and a velocity decoder and
applying the alignment loss to the encoder output.
VA-VAE \citep{yao2025vavae} moves the alignment out of the generator into the latent
space.
REPA-E \citep{leng2025repae} makes joint optimization of VAE and diffusion backbone possible. What is common in all these works is discarding the projection head after training, which, as we argue, is not optimal.
Another line of works relies on the supervisory signal from the model itself rather than external vision encoder. SRA \citep{jiang2026sra} distills the
representation of a later layer at low noise into an earlier layer at high
noise. Self-Flow \citep{chefer2026selfflow} instead creates information asymmetry
across tokens by assigning them different noise levels, so the model has to
infer corrupted tokens from cleaner ones. We provide a more detailed related-work discussion in Appendix~\ref{app:related}.

\textbf{Relation to REG.} REG \citep{wu2025reg} adds a
representation token and diffuses it jointly with the patch tokens, so the
token is generated from noise alongside the image while the alignment
projector is still discarded. REG performs best when aligning
with the encoder \textit{class} token. This is a limitation because not every vision encoder has the class token. In comparison, RECON does not have this requirement and thus can work well with various vision encoders. Moreover, RECON does not add new variables to be diffused.

\section{Preliminaries}
\label{sec:prelim}

\textbf{Flow matching.} Let $x_0$ be a clean image in RGB space. As in
latent diffusion \citep{rombach2022ldm}, we work with its latent
$z_0 = E_{\text{VAE}}(x_0)$, where $E_{\text{VAE}}$ is a pretrained VAE
encoder. We write $z_0$ as $N$ patches of dimension $d_z$, so it lies in
$\mathcal{Z} = \mathbb{R}^{N \times d_z}$. Flow matching
\citep{lipman2023flow} trains a network to transport noise
$z_1 \sim \mathcal{N}(0, I)$ of the same shape to such latents. Following
SiT \citep{ma2024sit}, the two are connected by the linear interpolant of
rectified flow \citep{liu2023rectifiedflow},
\begin{equation}
z_t = (1-t)\,z_0 + t\,z_1, \qquad t \in [0, 1],
\label{eq:interp}
\end{equation}
where $t=0$ is the clean latent and $t=1$ is pure noise. The velocity
along this path is $\mathrm{d}z_t/\mathrm{d}t = z_1 - z_0$, and a network
$v_\theta$ learns to predict it from $z_t$ and $t$ alone:
\begin{equation}
\mathcal{L}_{\text{FM}}(\theta) = \mathbb{E}_{t, x_0, z_1}
\Big[\, \lVert v_\theta(z_t, t) - (z_1 - z_0) \rVert^2 \,\Big].
\label{eq:fm-loss}
\end{equation}
SiT also derives from the same $v_\theta$ an
SDE with the same marginals, which we use in our experiments
(Section~\ref{sec:setup}). Diffusion models fit the same template with a
different interpolant and prediction target, so, without loss of generality,
we use flow-matching notation throughout.

\textbf{Intermediate representations in diffusion transformers.} We
parameterize $v_\theta$ with a diffusion transformer, as in SiT
\citep{ma2024sit}, which takes $z_t$ as a sequence of $N$ \emph{patch tokens}
of hidden dimension $C$ and processes it with a stack of $L$ transformer
blocks. Writing
$B_j$ for block $j$ and $h^{(j)} \in \mathcal{H} = \mathbb{R}^{N\times C}$
for its output,
\begin{equation}
h^{(j)} = B_j\big(h^{(j-1)},t\big), \qquad j=1,\ldots,L,
\label{eq:block-sequence}
\end{equation}
where $h^{(0)}$ holds the patch tokens of $z_t$ and we omit the class label
for brevity.

Following REPA \citep{yu2025repa}, one can view the diffusion model
$v_\theta(z_t, t)$ as a composition of two functions
$\mathcal{D}_\theta \circ \mathcal{E}_\theta$: an encoder
$\mathcal{E}_\theta : \mathcal{Z} \to \mathcal{H}$ with
$\mathcal{E}_\theta(z_t, t) = h^{(\ell)} = h$, made of the first $\ell$
blocks, and a decoder $\mathcal{D}_\theta : \mathcal{H} \to \mathcal{Z}$
with $\mathcal{D}_\theta(h, t) = \hat v$, made of the remaining $L-\ell$
blocks. We call these blocks the \emph{early} and the
\emph{later layers}, respectively. The two are parts of a jointly trained
transformer; the split marks where REPA supervises
the backbone and where RECON inserts its semantic tokens
(Fig.~\figpart{fig:teaser}{b,c}).

\textbf{Representation alignment.} Trained with the flow-matching loss
(Eq.~\ref{eq:fm-loss}) alone, the diffusion transformer learns semantic features
slowly \citep{yu2025repa}. REPA \citep{yu2025repa} speeds this up by
supervising $h$ with a frozen pretrained vision encoder $f_{\text{ext}}$,
such as DINOv2 \citep{oquab2024dinov2}. Let
$y_0 = f_{\text{ext}}(x_0) \in \mathbb{R}^{N \times D}$. A small trainable MLP, $g_\phi$, maps each token of $h$ into this feature space, shown in Fig.~\figpart{fig:teaser}{a}. REPA then maximizes the similarity between
$g_\phi(h)$ and $y_0$, patch by patch:
\begin{equation}
\mathcal{L}_{\text{align}}(\theta, \phi) = -\, \mathbb{E}_{t, x_0, z_1}
\Big[\, \frac{1}{N} \sum_{n=1}^{N}
\mathrm{sim}\big(y_0^{[n]},\; g_\phi(h^{[n]})\big) \,\Big],
\label{eq:repa-loss}
\end{equation}
where $n$ indexes patches, and $\mathrm{sim}$ is the cosine similarity. This
loss is added to Eq.~\ref{eq:fm-loss} with a weight $\lambda$.  We call $\ell$ the \emph{alignment depth} ($\ell = 8$ for SiT-XL/2
in REPA). After training, $g_\phi$ is discarded and only the diffusion transformer
is used for sampling.

\section{Motivation}
\label{sec:motivation}

We are motivated by the following three considerations.

\textbf{First}, {stronger alignment with external vision encoder \textit{positively correlates} with better generation}. For different REPA checkpoints, we plot their CKNNA and generation FID (gFID) in Fig.~\figpart{fig:alignment}{a}. CKNNA compares the nearest-neighbor structure of two representations; a higher CKNNA score means higher alignment. The figure demonstrates that alignment and generation improve together.

\textbf{Second}, features of $g_\phi$ have the \textit{highest similarity} with features of the external vision encoder. This is plotted in Fig.~\figpart{fig:alignment}{b}, where CKNNA of $g_\phi$ is clearly higher than all the transformer layers. This is understandable because $g_\phi$ is the mapping layer and stays closest to the vision encoder.

\textbf{Third}, existing practices \textit{discard} $g_\phi$ after training. As a result, the later layers only see $h$, which is less aligned with the vision encoder than the output of $g_\phi$ (Fig.~\figpart{fig:alignment}{b}).

Based on the above, we argue that it is not reasonable to simply drop the projector $g_\phi$, the most semantic-rich module in the diffusion backbone; we keep $g_\phi$ and give $\mathcal{D}_\theta$ direct access to the best-aligned features, at no extra cost.

\section{Proposed Representation Conditioning (RECON)}
\label{sec:method}

\textbf{Overview.} The key is to retain $g_\phi(h)$. At inference, we fuse the output of $g_\phi(h)$ with the diffusion patch tokens, which serves as in-context conditioning for the later diffusion layers. Before fusion, the $g_\phi(h)$ output is pooled into a few \textit{semantic tokens}. During training, we directly use the vision encoder features and pool them into semantic tokens, called \textit{teacher forcing}, because the projection layer is being trained and thus not stable. Using perception model features for training and projected features for inference creates a distribution gap. We mitigate this through a \textit{time-dependent scheduling for semantic token injection}. Overall framework is depicted in Fig.~\figpart{fig:teaser}{b} and \figpart{fig:teaser}{c}. We detail these components below.

\begin{figure}[t]
\centering
\includegraphics[width=\linewidth]{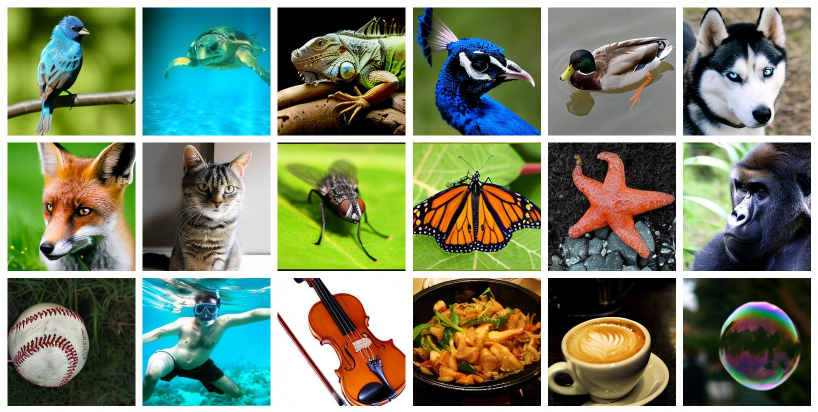}
\caption{\textbf{Qualitative results on ImageNet $256\times256$} from
SiT-XL/2 + RECON, using classifier-free guidance with $w=4$.}
\label{fig:qualitative}
\end{figure}

\subsection{Training pipeline}
\label{sec:training}

\textbf{Mapping vision encoder tokens to semantic tokens.}
Let $y_0 = f_{\text{ext}}(x_0) \in
\mathbb{R}^{N \times D}$ be the vision encoder features of the clean image. We pool them into $k \ll N$
semantic tokens and map these tokens to the hidden dimension of the model:
\begin{equation}
s = m\big(\mathrm{Pool}(y_0)\big) \in \mathbb{R}^{k \times C},
\label{eq:pool}
\end{equation}
where $C$ is the working dimension of transformer blocks. $\mathrm{Pool}$ is parameter-free and will produce a square number of semantic tokens, \textit{e.g.}, $k=1^2,2^2$. We adopt  $k=1$ by default. $m$ is a function that matches the dimensions. $m$ is the identity mapping when $D=C$; it is zero-padding if $D<C$; otherwise it is a linear mapping.
By default we use zero-padding, because the vision encoder dimension $D$ is usually lower than $C$ when the diffusion transformer is large enough, i.e., of size Base, Large, and Extra Large. Zero-padding adds no parameters. If we use a Small transformer, $D>C$, then we use a linear mapping $m$, which adds very small number of parameters, \emph{e.g.}, less than 1M parameters.  We study the choice of $k$ and $m$ in Table~\ref{tab:ablation}.

\textbf{In-context conditioning.} The semantic tokens are concatenated with patch tokens $h$. They are then inserted in the later blocks $\mathcal{D}_\theta$:
\begin{equation}
\tilde{h} = \big[\, h \;;\; s \,\big] \in \mathbb{R}^{(N+k) \times C},
\qquad \hat{v} = \mathcal{D}_\theta(\tilde{h}, t).
\label{eq:concat}
\end{equation}
In our method, every block of $\mathcal{D}_\theta$ processes patch and semantic tokens
together. The patch tokens attend to the semantic tokens at any
depth.
The semantic tokens are dropped before the final block.

\textbf{Teacher forcing.}
During training, we build semantic tokens from the vision encoder
features $y_0$ (Eq.~\ref{eq:pool}), because we have access to the clean image. This is the same teacher forcing method as language models: the
model is conditioned on the ground truth tokens rather than on its own
predictions. We could also use $g_\phi$ output to obtain semantic tokens. But we choose teacher forcing because $g_\phi$ is being trained.
Its output starts far from $y_0$ and keeps changing as training proceeds. Therefore,
conditioning on the $g_\phi$ output would ask the later layers to rely on a signal that
is neither accurate nor stable, whereas $y_0$ is stable from the first
training step. We compare teacher forcing with other training strategies in Table~\ref{tab:tokens}.

\textbf{Time-dependent scheduling for semantic token injection.} Teacher forcing at every timestep may
introduce a train--inference gap. At inference the semantic tokens come
from $g_\phi(h)$, which matches $y_0$ relatively well at low noise levels and poorly
at high noise levels. Training with $y_0$ at high $t$ (noisy steps) would therefore teach
$\mathcal{D}_\theta$ to rely on information that $g_\phi(h)$ cannot provide
at generation time. We mitigate this by injecting semantic tokens at timesteps only below
a threshold $t_{\max}$,
\begin{equation}
s =
\begin{cases}
m\big(\mathrm{Pool}(y_0)\big), & t < t_{\max} \\[2pt]
\varnothing, & t \geq t_{\max}
\end{cases},
\label{eq:schedule}
\end{equation}
where $\varnothing$ denotes that no semantic tokens are appended and
$\mathcal{D}_\theta$ runs on $h$ alone.

\input{tables/nocfg_system}

\subsection{Inference}
In inference, we use the output of $g_\phi$ in place of $y_0$ to obtain semantic tokens:
\begin{equation}
s = m\big(\mathrm{Pool}(g_\phi(h))\big).
\label{eq:infer-tokens}
\end{equation}
We use $g_\phi$ because 1) $g_\phi$ is trained to predict the vision-encoder features from $h$ and thus is a good proxy of the vision encoder and 2) there is no clean image available during inference. In the same way as training, the semantic tokens are then concatenated to the patch tokens and serve as in-context conditioning for later layers. There is no time-dependent scheduling for inference, the benefit of which is shown in Table~\ref{tab:inference-injection}.

\section{Experiments}
\label{sec:experiments}

We evaluate RECON introduced in
Section~\ref{sec:method}. Three questions organize the evaluation:

\begin{enumerate}[leftmargin=*]
\setlength{\itemsep}{1pt}\setlength{\parskip}{0pt}\setlength{\topsep}{2pt}
\item Does RECON improve generation
quality and accelerate convergence, for both class-conditional and text-to-image generation?
(Sections~\ref{sec:speed} and~\ref{sec:t2i})
\item How much does RECON improve the semantics of intermediate diffusion features? What is its benefit apart from generation quality? (Section~\ref{sec:dynamics})
\item Does the improvement hold across model scales, latent spaces, and target
encoders?
(Section~\ref{sec:generalization})
\end{enumerate}

\subsection{Setup}
\label{sec:setup}

\textbf{Implementation details.} We follow the setup of SiT \citep{ma2024sit}
and REPA \citep{yu2025repa} unless stated otherwise. We use ImageNet
\citep{deng2009imagenet} at $256\times256$ resolution and follow ADM
\citep{dhariwal2021adm} for data preprocessing. Each image is encoded into a
$32\times32\times4$ latent by the SD-VAE \citep{rombach2022ldm}. We use the
B/2, L/2 and XL/2 models of SiT, which have a patch size of 2, and a batch
size of 256. RECON uses alignment depth 8, DINOv2-B \mbox{\citep{oquab2024dinov2}} as vision encoder target, $k=1$, $t_{\max}=0.5$, and zero-padding for dimension matching (Table~\ref{tab:ablation}). For
text-to-image generation, we train MMDiT \citep{esser2024sd3} on MS-COCO
\mbox{\citep{lin2014coco}} and NanoGen \citep{leng2026diffusionbench}\footnote{NanoGen’s text-to-image setup, which uses a DDT backbone.} on BLIP3-o
\citep{chen2025blip3o} (Appendix~\ref{app:impl-t2i}). More details are in
Appendix~\ref{app:impl}.

\textbf{Datasets and evaluation protocols.} We report gFID, inception score, precision and
recall over 50K samples, drawn with an SDE sampler at 250 steps for ImageNet.  For text-to-image generation, we report FID and CLIP score on MS-COCO (83K images), and FID and IS on BLIP3o-256 (36M images). Samples are drawn with an ODE sampler at 50 steps.

\subsection{Class-conditional generation}
\label{sec:speed}

\textbf{C2I: comparison with existing methods.} Table~\ref{tab:nocfg} summarizes the comparisons using SD-VAE and without CFG. At 400K training steps, our method is stronger than state-of-the-art methods such as REG (3.40) and REPA-E (4.07). Longer training to 1M and 2.4M steps reduces gFID to 2.54 and 2.21, respectively. Fig.~\ref{fig:qualitative} shows samples from the final model. Fig.~\figpart{fig:teaser}{d} shows the training speedup.

\textbf{C2I: system-level comparisons.} Following RAE \citep{zheng2025rae}, we use Autoguidance \citep{karras2024autoguidance} for the guided results (more details in Appendix~\ref{app:impl}). Table~\ref{tab:system} summarizes the comparisons. At 80 epochs, RECON reaches 2.09, better than MaskDiT (2.28) and DiT (2.27), which are trained for 1600 and 1400 epochs, respectively. At 480 epochs, RECON reaches 1.49, better than SiT (2.06) and on par with REPA with scheduled CFG (1.42), which are trained for 1400 and 800 epochs, respectively. Similar to RAE, our preliminary observation is that CFG helps RECON less than it helps standard diffusion transformers. Designing guidance for semantically rich diffusion backbones, such as RECON and RAE, is a promising future direction, which we discuss in Sec.~\ref{sec:discussions}.

\subsection{Text-to-image generation}
\label{sec:t2i}

\begin{figure}[!ht]
\centering
\captionsetup{skip=4pt}
\includegraphics{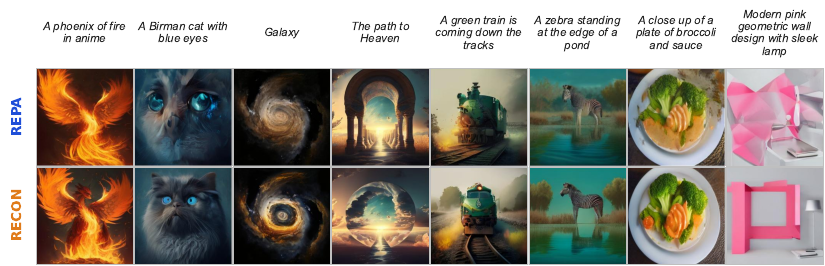}
\caption{\textbf{Qualitative comparison on text-to-image generation
(BLIP3-o).} Curated qualitative samples from REPA and RECON trained for
100K iterations. Prompts are shown on top.}%
\label{fig:t2isamples}
\end{figure}

\textbf{T2I: comparison with REPA.} We compare our method with REPA in Table~\ref{tab:coco} and Table~\ref{tab:blip3o}, where we report consistently lower FID, and higher CLIP score and IS. On two different datasets, the gFID of RECON is 8.26 and 3.90, while gFID of REPA is 9.85 and 5.41, respectively. The BLIP-3o results are more useful because the training set is much larger than MS-COCO. Qualitative results are shown in Fig.~\ref{fig:t2isamples}. These results indicate that RECON is effective for T2I training.

\input{tables/t2i_row}

\subsection{Effect on representation dynamics}
\label{sec:dynamics}

Beyond generation quality, we examine how RECON changes the semantics of
the intermediate features and how a sample forms during generation.

\needspace{13\baselineskip}

\begin{wrapfigure}{r}{0.355\textwidth}
\vspace{-\intextsep}
\centering
\includegraphics[scale=0.872,trim=2.484 2.880 1.976 1.976,clip]{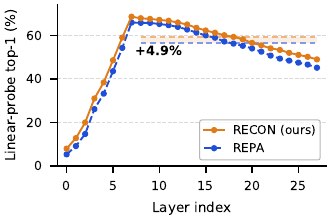}
\caption{\textbf{Linear probing vs.\ layers.} ImageNet-1K validation
top-1, at $t=0.25$.}
\label{fig:dynamics_lp}
\vspace{-0.5\intextsep}
\end{wrapfigure}
\textbf{RECON improves semantics in each diffusion layer.}
Fig.~\ref{fig:dynamics_cknna}
and Fig.~\ref{fig:dynamics_lp} use CKNNA and linear probing accuracy to measure semantics in each layer\footnote{For a fair comparison, we exclude the semantic tokens when evaluating both metrics.}, comparing REPA and RECON. We clearly observe that under both measures, RECON brings more semantics to each layer. Specifically, in the later layers that receive semantic tokens, RECON raises mean CKNNA by $21.2\%$ and mean linear probing accuracy by $4.9\%$.

\begin{figure}[!h]
\centering
\captionsetup[subfigure]{font=normalsize,justification=centering,skip=2pt}
\begin{subfigure}[b]{252pt}
\centering
\includegraphics[width=\linewidth]{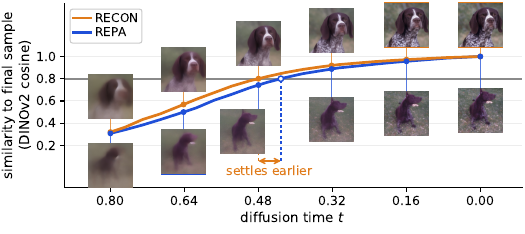}
\subcaption{Semantic emergence over generation}
\label{fig:settle_traj}
\end{subfigure}\hfill
\begin{subfigure}[b]{137pt}
\centering
\includegraphics[width=\linewidth]{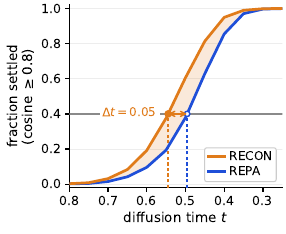}
\subcaption{When does structure settle?}
\label{fig:settle_cdf}
\end{subfigure}
\caption{\textbf{RECON settles the structure of a sample earlier in
generation.} (a) measures how close the running estimate $\hat{x}_0(t)$
is, in semantic space, to the image that the same run finally produces. It
is measured for a single sample, where RECON is above REPA at every step
(both methods start from the same noise). In (b), we generate 10K images. At
each diffusion timestep, we measure the percentage of samples whose running
estimate $\hat{x}_0(t)$ similarity with the final image is greater than
0.8. RECON is above REPA.}
\label{fig:settle}
\end{figure}

\textbf{Semantic structure emerges earlier in sampling.} The richer
intermediate features also change how a sample develops over generation. We
visualize the \emph{semantic trajectory} of a generated sample: at
each step we measure the cosine similarity between the running estimate $\hat{x}_0(t)$ and the final image, in the space of DINOv2. Fig.~\ref{fig:settle} presents how much of the image subject and layout is already decided at time $t$. Fig.~\figpart{fig:settle}{a} shows one sample, while Fig.~\figpart{fig:settle}{b} shows the statistics of $10K$ samples.
We clearly observe that RECON allows such layout to appear earlier than REPA.
A similar behavior is observed under perturbation:
given a heavily corrupted image, RECON recovers the identity and overall
composition better than REPA (Fig.~\ref{fig:sdedit}).

\begin{wraptable}{r}{0.36\textwidth}
\vspace{-\intextsep}
\centering
\captionsetup{font=small,skip=3pt}
\caption{\textbf{Sampling steps.} gFID-50K; SiT-XL/2, 400K iter., no CFG.}
\label{tab:steps}
\footnotesize
\setlength{\tabcolsep}{5pt}
\begin{tabular}{lccc}
\toprule
NFE & 50 & 100 & 250 \\
\midrule
REPA & 10.8 & 8.4 & 7.9 \\
\rowcolor{gray!12}
RECON & 3.3 & 3.2 & 3.2 \\
\bottomrule
\end{tabular}
\vspace{-0.5\intextsep}
\end{wraptable}
\textbf{RECON works with fewer sampling steps.} Since the structure of samples emerges earlier, RECON needs fewer sampling steps for generation. In Table~\ref{tab:steps}, we reduce NFE to 100 and 50 and keep everything else
fixed. gFID of REPA degrades from 7.9 to 8.4 and 10.8, respectively. In comparison,
RECON keeps gFID of 3.2 at 100 steps, and goes to 3.3 at 50 steps. This
means 2.5$\times$ to 5$\times$ fewer network evaluations at inference, with almost no cost in generation.

\subsection{Generalization and scalability}
\label{sec:generalization}

Unless stated otherwise every run in this section uses SD-VAE latents, DINOv2-B as the target encoder and alignment depth $8$, and is evaluated without classifier-free guidance.
Every baseline is vanilla REPA trained under the same protocol.
The full metrics are in Appendix~\ref{app:extra}.

\begin{figure}[t]
\centering
\captionsetup[subfigure]{font=normalsize,justification=centering,skip=0pt}
\raisebox{-5.97pt}{\includegraphics[scale=0.794,trim=1.710 2.862 1.076 1.076,clip]{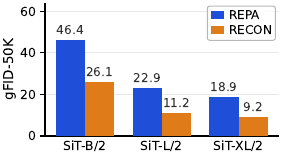}}\hfill
\raisebox{-5.60pt}{\includegraphics[scale=0.794,trim=1.620 3.330 1.076 1.076,clip]{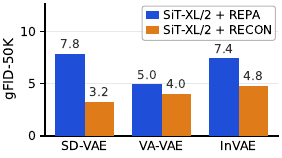}}\hfill
\raisebox{-10.94pt}{\includegraphics[scale=0.794,trim=1.710 3.798 1.076 1.076,clip]{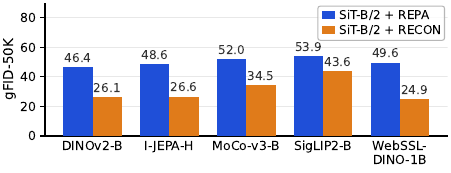}}\par\vspace{3pt}
\begin{subfigure}[t]{105.1pt}\subcaption{Variation in model scale}\label{fig:genscale}\end{subfigure}\hfill
\begin{subfigure}[t]{105.2pt}\subcaption{Variation in the tokenizers}\label{fig:genlatents}\end{subfigure}\hfill
\begin{subfigure}[t]{169.5pt}\subcaption{Variation in the target encoders}\label{fig:genencoder}\end{subfigure}
\caption{\textbf{RECON \textit{vs}. REPA across model scale, tokenizer and target encoder.}
gFID-50K of REPA and RECON. (a) SiT-B/2, SiT-L/2 and SiT-XL/2 at 20 epochs.
(b) Three tokenizers at 80 epochs. (c) Five target vision encoders, SiT-B/2 at
20 epochs.}
\label{fig:generalization}
\end{figure}

\textbf{Impact of model size.} Fig.~\ref{fig:genscale} compares SiT-B/2,
SiT-L/2 and SiT-XL/2 at 20 epochs. RECON improves REPA significantly at every model scale.
Interestingly, SiT-L/2 with RECON
reaches gFID = $11.24$, which is better than $18.88$ for SiT-XL/2 with REPA.

\textbf{Effectiveness of RECON for different VAEs.} We compare RECON and REPA on three VAEs: SD-VAE \citep{rombach2022ldm}, VA-VAE \citep{yao2025vavae} and InVAE \citep{leng2025repae}. Results are in Fig.~\ref{fig:genlatents}.
RECON yields consistent improvements over REPA for each tokenizer: gFID improves
from $7.84 \to 3.25$ on SD-VAE, from $7.41 \to 4.78$ on InVAE, and from
$4.95 \to 4.02$ on VA-VAE. The improvement on VA-VAE is smaller because it is trained with an additional REPA loss, so a part of the semantics RECON tries to inject into the diffusion backbone is already available. For further discussions see Appendix \ref{app:discussion}.

\input{tables/ablation_design}

\textbf{Effectiveness of RECON for different target vision encoders.} We use five encoders: I-JEPA-H \citep{assran2023ijepa}, DINOv2-B, MoCo-v3-B \citep{chen2021mocov3}, SigLIP2-B \citep{tschannen2025siglip2}, and WebSSL-DINO-1B \citep{fan2025webssl} and summarize results in Fig.~\ref{fig:genencoder}. Across the encoders, RECON also gives consistent gFID improvement over REPA.

\textbf{Ablations and variations on design components.} Table~\ref{tab:ablation} varies the
components RECON introduces, one block at a time: the alignment depth, the number of semantic
tokens, the injection threshold $t_{\max}$, and the dimension-matching
mechanism. For \bm{\hlwrap{ctmax}{$t_{\max}$}}, we choose 0.5. Larger $t_{\max}$ makes diffusion exposed to unseen semantics during inference; smaller $t_{\max}$ makes it harder for diffusion blocks to understand how patch-tokens should interact with semantics. We choose \bm{\hlwrap{ctok}{$k$}} to be 1. If we increase $k$, errors in $g_\phi$ prediction are amplified and can harm the generation quality. For the \bm{\hlwrap{cdm}{dimension-matching mechanism}}, we choose zero-padding to increase the dimensions of semantic tokens to be the same as patch tokens. This method is easier and appears to be more effective than linear mapping. \bm{\hlwrap{cdepth}{Depth of alignment}} analysis suggests that for XL models, Layer 8 is a good choice for alignment. It is also worth mentioning that RECON performance is relatively stable within a reasonable range of this factor.

\section{Discussions}
\label{sec:discussions}

\textbf{Why do later layers need more semantics?} Later layers are only supervised by the flow-matching loss and only depend on $h$. In comparison, the gradient of REPA loss is back-propagated to early layers, so early layers are optimized towards higher semantics. By injecting semantics to late layers, the late layers, through training, can decide whether to use them or not, which is at least not detrimental and turns out to be beneficial.

\textbf{Rethinking guidance for semantically rich diffusion models.} CFG helps RECON less than it helps standard diffusion transformers. Our hunch is that with the semantic tokens, the backbone is already rich in semantics without the class label, so the unconditional prediction is close to the conditional one. RAE \citep{zheng2025rae} reports a similar observation, and like RAE, we use Autoguidance \citep{karras2024autoguidance}. We further improve Autoguidance with two simple changes at sampling time, \textcolor{cmethod}{\emph{Projected Autoguidance}} (Appendix~\ref{app:pag}) and \textcolor{cmethod}{\emph{Semantic Token Extrapolation}} (Appendix~\ref{app:extrap}). They improve gFID from $2.26 \to 2.09$ at 80 epochs and from $1.80 \to 1.49$ at 480 epochs. We see these as first steps. One promising direction is Internal Guidance \citep{zhou2026internal}, which uses an intermediate layer of the model as the weak model. Another is guidance that acts directly on the semantic tokens, rather than on the velocity.

\textbf{Limitations.} RECON reuses a projection head that a REPA-style alignment loss
has already trained, so it cannot be applied to a model trained without such a loss. What it removes
is the waste in discarding that head, not the need for alignment in the first place. More in Appendix~\ref{app:limitations}.

\section{Conclusion}
\label{sec:conclusion}
This paper challenges the practice in representation alignment methods that the projection head $g_\phi$ should be discarded after training. This is based on our observation that the output of $g_\phi$ is very semantical: it aligns with the vision encoder more strongly than any of the diffusion transformer layers. Discarding it means we forfeit useful and also free cues. We propose representation conditioning (RECON), which supplies output of $g_\phi$ to the later layers. With no added parameters and no change in training loss, RECON simply lets later layers decide how to use the conditioning semantics. We report good improvement over REPA in various settings. We hope that RECON was effective in conveying and bringing up the following message: \textit{what matters in representation alignment is not only acquiring a representation, but also keeping it accessible in generation.}

\subsection*{AI use statement}
In this work, we used Claude Code~\citep{anthropic2026claude} to assist with implementing our methods, preparing experiment configurations, and running experiments. Additionally, we used Claude Code to draft the paper outline, check the grammar of the manuscript, and improve the clarity of the writing. We also used it to generate most of the figures and tables from our experiment logs and checkpoints, following the colour palette we provided. We have reviewed all AI-assisted work, including the code, experiment configurations, figures, tables, and text. We take responsibility for the final content of this work and the associated codebase, including all materials produced with AI assistance.

\subsection*{Reproducibility statement}
To support reproducibility, we provide all implementation details and hyperparameters in Appendix~\ref{app:impl}.

\clearpage
\setlength{\bibsep}{5pt}
\bibliography{references}
\bibliographystyle{recon}

\appendix

\section{Extended related work}
\label{app:related}

Section~\ref{sec:related} covers representation alignment. Here we extend
it with the diffusion backbones we compare against, other ways of bringing
pretrained representations into diffusion models, and train--inference gaps.

\textbf{Diffusion transformers.} DiT \citep{peebles2023dit} replaced the
U-Net of latent diffusion \citep{rombach2022ldm} with a transformer and
showed that generation quality improves with model compute. SiT
\citep{ma2024sit} keeps the DiT architecture and trains it with the
interpolant framework, which is the setting we follow. Other works change
the architecture or add objectives during training: U-ViT
\citep{bao2023uvit} adds long skip connections between shallow and deep
blocks, DiffiT \citep{hatamizadeh2024diffit} makes self-attention
time-dependent, and MDT \citep{gao2023mdt}, MaskDiT \citep{zheng2024maskdit}
and SD-DiT \citep{zhu2024sddit} add masking or self-supervised objectives.
For text-to-image generation, MMDiT \citep{esser2024sd3} keeps text and
image tokens in separate streams that interact through joint attention.

\textbf{Pretrained representations in the latent space.} Instead of
aligning the generator, some works build the semantics into the latent
space itself. VA-VAE and REPA-E (Section~\ref{sec:related}) align or
fine-tune the VAE. RAE \citep{zheng2025rae} goes further and replaces the
VAE encoder with a frozen pretrained representation encoder such as DINOv2,
training only a decoder and running diffusion in that representation space.
These approaches are complementary to RECON: in our tokenizer experiments
(Fig.~\ref{fig:genlatents}), RECON still improves generation on VA-VAE,
although by a smaller margin than on SD-VAE.

\textbf{Train--inference gaps and forcing strategies.} Conditioning a model
on information during training that is not available at inference is a
well-known problem. In autoregressive video diffusion, teacher forcing conditions on context
that the model must later produce itself, and Self Forcing
\citep{huang2025selfforcing} closes the resulting gap by rolling out the
model during training. RECON has the same kind of gap: it uses teacher
forcing, conditioning on vision encoder features during training and on the
output of $g_\phi$ at inference. We reduce this gap with time-dependent
scheduling for semantic token injection (Section~\ref{sec:method}).

\section{Hyperparameter and more implementation details}
\label{app:impl}

\providecommand{\chk}[1]{\textcolor{red}{#1}}

\begin{table}[h]
\footnotesize
\setlength{\tabcolsep}{3.5pt}
\captionsetup{justification=raggedright,singlelinecheck=false}
\begin{minipage}[t]{0.485\textwidth}
\centering
\setlength{\tabcolsep}{3pt}%
\caption{\textbf{Hyperparameters, class-conditional.} ImageNet
$256\times256$.}
\label{tab:hparams}
\begin{tabular}{lccc}
\toprule
 & SiT-B/2 & SiT-L/2 & SiT-XL/2 \\
\midrule
\multicolumn{4}{l}{\emph{Architecture}} \\
Input dim. & \multicolumn{3}{c}{$32\times32\times4$} \\
Num. layers & 12 & 24 & 28 \\
Hidden dim. & 768 & 1024 & 1152 \\
Num. heads & 12 & 16 & 16 \\
\midrule
\multicolumn{4}{l}{\emph{Alignment (REPA)}} \\
$\lambda$ & \multicolumn{3}{c}{1.5} \\
Alignment depth & 4 & 8 & 8 \\
Similarity & \multicolumn{3}{c}{cos.\ sim.} \\
Vision encoder & \multicolumn{3}{c}{DINOv2-B} \\
\midrule
\multicolumn{4}{l}{\emph{RECON}} \\
Sem.\ tokens $k$ & \multicolumn{3}{c}{1} \\
$t_{\max}$ (training) & \multicolumn{3}{c}{0.5} \\
Dim.\ matching & identity & zero-pad & zero-pad \\
\midrule
\multicolumn{4}{l}{\emph{Optimization}} \\
Iterations & 100K & 100K & 400K--2.4M \\
Batch size & \multicolumn{3}{c}{256} \\
Optimizer & \multicolumn{3}{c}{AdamW} \\
Learning rate & \multicolumn{3}{c}{$10^{-4}$} \\
$(\beta_1, \beta_2)$ & \multicolumn{3}{c}{(0.9, 0.999)} \\
\midrule
\multicolumn{4}{l}{\emph{Interpolant and sampling}} \\
Objective & \multicolumn{3}{c}{$v$-prediction} \\
Sampler & \multicolumn{3}{c}{Euler--Maruyama} \\
Sampling steps & \multicolumn{3}{c}{250} \\
Guidance$^\ddagger$ & \multicolumn{3}{c}{Autoguidance} \\
Weak model & \multicolumn{3}{c}{SiT-XL/2 + RECON, 125K iter.} \\
Guidance scale $w$ & \multicolumn{3}{c}{1.4} \\
Guidance interval & \multicolumn{3}{c}{$(0, 0.85)$} \\
\bottomrule
\multicolumn{4}{l}{\scriptsize $^\ddagger$Only for the guided SiT-XL/2 results.}
\end{tabular}
\end{minipage}\hfill
\begin{minipage}[t]{0.485\textwidth}
\centering
\caption{\textbf{Hyperparameters, text-to-image.} Settings not listed
are as in Table~\ref{tab:hparams}.}
\label{tab:hparams-t2i}
\begin{tabular}{lcc}
\toprule
 & MS-COCO & BLIP3-o \\
\midrule
\multicolumn{3}{l}{\emph{Architecture}} \\
Backbone & MMDiT & NanoGen \\
Num. layers$^\dagger$ & 24 & 28 + 2 \\
Hidden dim.$^\dagger$ & 768 & 1152 / 2048 \\
Num. heads & 24 & 16 \\
\midrule
\multicolumn{3}{l}{\emph{Text conditioning}} \\
Text encoder & CLIP ViT-L/14 & Qwen3-0.6B \\
Max. text tokens & 77 & 256 \\
Timestep tokens & 0 (adaLN) & 4 \\
CFG dropout & 0.1 & 0.1 \\
\midrule
\multicolumn{3}{l}{\emph{Alignment (REPA)}} \\
$\lambda$ & 1.5 & 1.5 \\
\midrule
\multicolumn{3}{l}{\emph{Optimization}} \\
Iterations & 150K & 100K \\
Batch size & 256 & 1024 \\
Optimizer & AdamW & AdamW \\
Learning rate & $10^{-4}$ & $2\times10^{-4}$ \\
$(\beta_1, \beta_2)$ & (0.9, 0.999) & (0.9, 0.95) \\
EMA decay & 0.9999 & 0.9995 \\
\midrule
\multicolumn{3}{l}{\emph{Interpolant and sampling}} \\
Time sampling & uniform & logit-normal \\
Sampler & ODE (Heun) & ODE (Euler) \\
Sampling steps & 50 & 50 \\
Guidance scale $w$ & 1, 2 & 1, 8 \\
\midrule
\multicolumn{3}{l}{\emph{Data}} \\
Training images & $\sim$83K & $\sim$36M \\
\bottomrule
\multicolumn{3}{l}{\scriptsize $^\dagger$NanoGen: encoder + decoder. (A text-to-image variant of DDT)}
\end{tabular}
\end{minipage}
\end{table}

\subsection{Class-conditional generation}
\label{app:impl-c2i}

\textbf{Further implementation details.} We build on the official REPA
implementation, which follows SiT \citep{ma2024sit}. We use AdamW with a
constant learning rate of $10^{-4}$, $(\beta_1,\beta_2)=(0.9,0.999)$ and
no weight decay, and train with mixed precision (fp16) and gradient
clipping. We pre-compute the SD-VAE latents and apply no data
augmentation. As in REPA, the projection head $g_\phi$ is a three-layer
MLP with SiLU activations. Table~\ref{tab:hparams} lists the full setup.

\textbf{RECON settings.} $\mathrm{Pool}$ in Eq.~\ref{eq:pool} is
average pooling of the vision encoder tokens to $k$ tokens. The
semantic tokens are concatenated with the patch tokens after the alignment
depth and dropped before the final block (Section~\ref{sec:method}). Let
$D$ be the feature dimension of the vision encoder and $C$ the hidden
dimension of the diffusion transformer. The dimension-matching function $m$
is the identity when $D=C$, zero-padding when $D<C$, and a linear layer
when $D>C$; for DINOv2-B ($D=768$), this gives the entries in
Table~\ref{tab:hparams}. During training, the tokens come from the vision
encoder and are injected only for $t<t_{\max}$. During inference, they come
from $g_\phi(h)$ and are injected at every step
(Table~\ref{tab:inference-injection}).

\textbf{Patch size per tokenizer.} For the tokenizer study
(Table~\ref{tab:latents}), we keep the XL backbone fixed and only change its
patch size, so that every run processes the same number of tokens. SD-VAE
\citep{rombach2022ldm} downsamples the $256\times256$ image by 8 and gives a
$32\times32\times4$ latent. We therefore use a patch size of 2 (SiT-XL/2).
VA-VAE \citep{yao2025vavae} and InVAE \citep{leng2025repae} downsample by 16
and give a $16\times16\times32$ latent. Following REPA-E, we use a patch size
of 1 for them (SiT-XL/1). In both cases, the transformer processes
$16\times16=256$ patch tokens, so the sequence length and compute per step are
the same across the three tokenizers.

\textbf{Pretrained encoders.} DINOv2-B \citep{oquab2024dinov2} is the
default target. For the encoder study (Fig.~\ref{fig:genencoder}), we also
use I-JEPA-H \citep{assran2023ijepa}, MoCo-v3-B \citep{chen2021mocov3},
SigLIP2-B \citep{tschannen2025siglip2} and WebSSL-DINO-1B
\citep{fan2025webssl}, all with their official checkpoints.

\textbf{Epochs and iterations.} ImageNet-1K has about 1.28M training
images, so at a batch size of 256 one epoch is about 5K iterations. Hence,
100K, 200K, 400K, 1M and 2.4M iterations correspond to 20, 40, 80, 200 and
480 epochs, respectively. All checkpoints are evaluated with the protocol of
Section~\ref{sec:setup}.

\textbf{Computing resources.} All models are trained on 4 NVIDIA H100 GPUs.
In our default setting (SiT-XL/2, batch size 256), training runs at about
4.10 iterations per second. A 400K-iteration run (80 epochs) therefore takes
about 27 hours (about 110 GPU-hours), and the longest 2.4M-iteration run
(480 epochs) about 6.8 days (about 650 GPU-hours). RECON adds only $k=1$
semantic token to the 256 patch tokens of the later layers, so its training
cost is almost the same as that of REPA.

\subsection{Text-to-image generation}
\label{app:impl-t2i}

Table~\ref{tab:hparams-t2i} lists the settings.

\textbf{MS-COCO.} We follow the text-to-image setup of U-ViT \citep{bao2023uvit}. We train MMDiT \citep{esser2024sd3} from scratch
on the training split of MS-COCO \citep{lin2014coco} and evaluate on the
validation split.

\textbf{BLIP3-o.} We use the NanoGen configuration of DiffusionBench
\citep{leng2026diffusionbench} and train on BLIP3-o \citep{chen2025blip3o},
about 36M image--text pairs at $256\times256$.

\section{Improving guided sampling}
\label{app:guidance}

Table~\ref{tab:system} reports guided results of RECON with Autoguidance
\citep{karras2024autoguidance} and, marked $\ddagger$, with two changes to
it, both at sampling time. \emph{Projected Autoguidance}
removes the part of the guidance update that only rescales the latent,
which is what over-saturates samples at large guidance scales.
\emph{Semantic Token Extrapolation} is specific to RECON: it applies the
idea of CFG to the predicted semantic token, with the token predicted at
$t=1$ in the role of the unconditional prediction. Neither change needs retraining. Algorithms~\ref{alg:pag}
and~\ref{alg:extrap} give the pseudocode.

\textbf{Notation.} As in Section~\ref{sec:prelim},
$z_t = (1-t)\,z_0 + t\,z_1$, where $t=0$ is the clean latent and $t=1$ is
pure noise, and the network predicts the velocity $v$. A velocity
prediction at $z_t$ gives the clean-latent estimate
\begin{equation}
\hat z_0 = z_t - t\,v.
\label{eq:z0-est}
\end{equation}
Autoguidance uses two models: the \emph{strong} model, which we sample
from, and a \emph{weak} model, an early checkpoint of the same model
(SiT-XL/2 + RECON at 125K iterations, Table~\ref{tab:hparams}). We write
$v_{\text{s}}$ and $v_{\text{w}}$ for their predictions, and
$\hat z_0^{\text{s}}$ and $\hat z_0^{\text{w}}$ for their clean estimates.

\subsection{Projected Autoguidance}
\label{app:pag}

\textbf{Autoguidance.} Autoguidance extrapolates from the weak model
towards the strong one:
\begin{equation}
v = v_{\text{w}} + w\,(v_{\text{s}} - v_{\text{w}})
  = v_{\text{s}} + (w-1)\,d,
\qquad d = v_{\text{s}} - v_{\text{w}},
\label{eq:ag}
\end{equation}
where $w$ is the guidance scale, and $w=1$ samples from the strong model
alone. Following \citet{kynkaanniemi2024interval}, we apply guidance only in
an interval of noise levels, $t \in (0, t_{\text{hi}})$, and use
$v_{\text{s}}$ alone outside it.

\textbf{Why large guidance scales over-saturate.} With Autoguidance,
samples become over-saturated once $w$ is above about 2. To see why, we
look at the update in terms of clean estimates. By Eq.~\ref{eq:z0-est},
Eq.~\ref{eq:ag} is the same as
\begin{equation}
\hat z_0 = \hat z_0^{\text{s}} + (w-1)\,\Delta,
\qquad \Delta = \hat z_0^{\text{s}} - \hat z_0^{\text{w}} = -t\,d.
\label{eq:ag-z0}
\end{equation}
At each spatial position, we
split $\Delta$ into a part parallel to $\hat z_0^{\text{s}}$ and a part
orthogonal to it. The parallel part does not change the direction of
$\hat z_0^{\text{s}}$; it only makes it longer or shorter. Yet it is large:
in our measurements, it makes up 50--60\% of the norm of $\Delta$. As $w$
grows, this rescaling grows with it, and the decoded image becomes
over-saturated. APG \citep{sadat2025apg} makes the same observation for CFG.

\textbf{Projection.} We remove the parallel part and keep the rest. At
every position,
\begin{equation}
\Delta_\perp = \Delta
  - \frac{\langle \Delta, \hat z_0^{\text{s}} \rangle}
         {\lVert \hat z_0^{\text{s}} \rVert^2}\, \hat z_0^{\text{s}},
\qquad
\hat z_0 = \hat z_0^{\text{s}} + (w-1)\,\Delta_\perp.
\end{equation}
Since $\Delta = -t\,d$, this is the same as projecting $d$, and in terms of
velocity it reads
\begin{equation}
v = v_{\text{s}} + (w-1)\,d_\perp,
\qquad
d_\perp = d
  - \frac{\langle d, \hat z_0^{\text{s}} \rangle}
         {\lVert \hat z_0^{\text{s}} \rVert^2}\, \hat z_0^{\text{s}}.
\label{eq:pag}
\end{equation}
Compared with Autoguidance (Eq.~\ref{eq:ag}), the only change is
$d \to d_\perp$ (Algorithm~\ref{alg:pag}). Projected Autoguidance works on
the latent directly. It needs no VAE decoding and no extra forward pass, so
it costs the same as Autoguidance. It also tolerates larger guidance
scales: we can use $w = 2.5$--$3$ before samples over-saturate, instead of
about 2.

\begin{algorithm}[!h]
\caption{Autoguidance vs.\ Projected Autoguidance}
\label{alg:pag}
\begin{minipage}[t]{0.48\linewidth}
\begin{lstlisting}[style=pseudo,emph={net_s,net_w}]
# (a) Autoguidance
# net_s, net_w: strong and weak models
# z: current samples at time t
# w: guidance scale

v_s = net_s(z, t)
v_w = net_w(z, t)
d = v_s - v_w

v = v_s + (w - 1) * d
\end{lstlisting}
\end{minipage}\hfill{\color{black!30}\vrule}\hfill
\begin{minipage}[t]{0.48\linewidth}
\begin{lstlisting}[style=pseudo,emph={net_s,net_w,dot}]
# (b) Projected Autoguidance
# dot: inner product over the latent
#   channels, at each spatial position

v_s = net_s(z, t)
v_w = net_w(z, t)
d = v_s - v_w

# clean estimate of the strong model
z0_s = z - t * v_s
# drop the part of d parallel to z0_s
c = dot(d, z0_s) / dot(z0_s, z0_s)
d_perp = d - c * z0_s

v = v_s + (w - 1) * d_perp
\end{lstlisting}
\end{minipage}
\end{algorithm}

\textbf{Relation to APG.} The projection comes from APG
\citep{sadat2025apg}, which splits the CFG update into parts parallel and
orthogonal to the conditional prediction. We differ from APG in three ways.
First, we apply the projection to Autoguidance instead of CFG. Second, we
project at each spatial position, while APG projects the whole latent at
once; per-position projection gave a lower FID in our experiments. Third, we
only drop the parallel part. APG also keeps a down-weighted parallel part
and adds rescaling and momentum, which we do not use.

\subsection{Semantic Token Extrapolation}
\label{app:extrap}

\textbf{Predicted semantic tokens across noise levels.} During training, the
semantic token comes from the DINOv2 features of one real image (teacher
forcing, Section~\ref{sec:training}). During sampling, it is predicted by
the projection head $g_\phi$ from the noisy input, and how specific this
prediction is depends on the noise level. At low noise levels, the semantics
of the image are easy to extract, so the predicted token is specific to the
sample. At high noise levels, it is hard to tell what the final image will
look like, so the predicted token is close to the average token of the
class and is less specific to the sample. As a result, at high noise
levels the predicted semantic token is more typical of its class than the token of
the final generated image.

\textbf{Extrapolating the token.} We take advantage of this with the same
idea as CFG, which extrapolates away from an unconditional prediction. Here,
the token predicted at $t=1$, $\hat u_1$, plays the role of the
unconditional prediction: at $t=1$ the input is pure noise, so $\hat u_1$
is close to the class average (cosine similarity 0.97 with the mean token
of the class). Let $\hat u_t$ be the semantic token predicted at time $t$. In the strong
model, we replace it with
\begin{equation}
u'_t = \hat u_1 + (1+\lambda)\,(\hat u_t - \hat u_1)
     = \hat u_t + \lambda\,(\hat u_t - \hat u_1),
\label{eq:extrap}
\end{equation}
with $\lambda = 0.05$ (Algorithm~\ref{alg:extrap}). Intuitively, $\hat u_t - \hat u_1$ is what the
token has learned about this particular sample beyond the class average, and
the extrapolation amplifies it. Since $\hat u_1$ comes from the model
itself, we do not need the class average from the training data. The weak
model is unchanged. Computing $\hat u_1$ needs only the first $\ell$ blocks
and $g_\phi$, once per sample, which costs about 0.3 of a forward pass.

\begin{algorithm}[!h]
\caption{Semantic Token Extrapolation in the strong model}
\label{alg:extrap}
\begin{minipage}[t]{0.48\linewidth}
\begin{lstlisting}[style=pseudo,emph={early,later,g,pool,m}]
# (a) RECON
# early: the first $\textcolor{codecomment}{\ell}$ blocks
# later: the remaining blocks
# g: projection head
# pool, m: pooling, dim. matching

h = early(z, t)
u_t = pool(g(h))

v_s = later(h, m(u_t), t)
\end{lstlisting}
\end{minipage}\hfill{\color{black!30}\vrule}\hfill
\begin{minipage}[t]{0.48\linewidth}
\begin{lstlisting}[style=pseudo,emph={early,later,g,pool,m}]
# (b) RECON + Token Extrapolation
# z_1: initial noise of this sample
# lam: extrapolation strength (0.05)

# once per sample, first $\textcolor{codecomment}{\ell}$ blocks only
u_1 = pool(g(early(z_1, 1)))

h = early(z, t)
u_t = pool(g(h))
u = u_t + lam * (u_t - u_1)
v_s = later(h, m(u), t)
\end{lstlisting}
\end{minipage}
\end{algorithm}

\section{Representation analysis}
\label{app:cknna}

\subsection{CKNNA}
Centered Kernel Nearest-Neighbor Alignment (CKNNA)
\citep{huh2024platonic} measures whether two representations give the same
\emph{local geometry} to the same set of images.  It is the local counterpart
of Centered Kernel Alignment (CKA) \citep{kornblith2019cka}.

Let $\phi_i\in\mathbb{R}^{d_\phi}$ and
$\psi_i\in\mathbb{R}^{d_\psi}$ be the two representations of image $i$, for
$i=1,\ldots,n$, and let $K_{ij}=\langle\phi_i,\phi_j\rangle$ and
$L_{ij}=\langle\psi_i,\psi_j\rangle$ be their Gram matrices.  CKA compares $K$
and $L$ over all image pairs with the Hilbert--Schmidt Independence Criterion
(HSIC).  CKNNA keeps only pairs that are neighbours in \emph{both} spaces.  If
$\mathcal{N}_{K}^{k}(i)$ and $\mathcal{N}_{L}^{k}(i)$ are the $k$ nearest
neighbours of image $i$ (excluding $i$) under $K$ and $L$, define
\begin{equation}
M_{ij}^{K,L}
=\mathbf{1}\!\left[j\in\mathcal{N}_{K}^{k}(i)\cap\mathcal{N}_{L}^{k}(i)\right],
\qquad
A_k(K,L)=\operatorname{HSIC}_u\!\left(M^{K,L}\odot K,\;M^{K,L}\odot L\right),
\end{equation}
where $\odot$ is the element-wise product and $\operatorname{HSIC}_u$ is the
unbiased estimator of \citet{song2012hsic}. For $\tilde K,\tilde L$ with zero
diagonals, as the masked kernels are,
\begin{equation}
\operatorname{HSIC}_u(\tilde K,\tilde L)
=\frac{1}{n(n-3)}\left[\operatorname{tr}(\tilde K\tilde L)
+\frac{\mathbf{1}^{\top}\tilde K\mathbf{1}\;\mathbf{1}^{\top}\tilde L\mathbf{1}}{(n-1)(n-2)}
-\frac{2}{n-2}\,\mathbf{1}^{\top}\tilde K\tilde L\mathbf{1}\right].
\end{equation}
The normalized score is
\begin{equation}
\operatorname{CKNNA}_k(K,L)
=\frac{A_k(K,L)}
{\sqrt{A_k(K,K)\,A_k(L,L)}}.
\end{equation}
Thus a high score means that images grouped together by one representation
are also grouped together by the other, with similar affinities.  Unlike a
direct feature distance, CKNNA remains meaningful when $d_\phi\ne d_\psi$, and
it does not change if either representation is rescaled.  With $k=n-1$, every
off-diagonal pair is kept and CKNNA reduces to CKA with the unbiased HSIC
estimator, whereas a small $k$ focuses the comparison on local semantic
structure.

For every CKNNA result in this paper, we randomly sample $n=10{,}000$
ImageNet validation images and use $k=10$, following REPA.  We spatially
average the dense patch features to obtain one vector per image, discard
RECON's semantic tokens before pooling, and $\ell_2$-normalize the vectors
before forming the linear kernels.  The
reference is the clean-image DINOv2-B representation; the layer-wise results
use the $400$K-iteration EMA checkpoint at $t=0.25$.

\subsection{Linear probing}
\label{app:linprobe}
We follow the linear probing protocol of MAE
\citep{he2022mae}. For each layer, we take the frozen features of the
400K-iteration EMA checkpoint at $t=0.25$. We use the null class
label, so that no label information leaks into the features. As for CKNNA,
we average the patch tokens to get one vector per image, and discard RECON's
semantic tokens before pooling. The classifier is a BatchNorm layer without
affine parameters, followed by a linear layer. We train it on the ImageNet-1K
training set and report top-1 accuracy on the validation set. Unlike the original
recipe, which trains for 90 epochs, we train the classifier for 20 epochs to
reduce the cost of probing every layer. REPA and RECON use the same schedule,
so the comparison between them is fair.

\section{Evaluation details}
\label{app:eval}

For ImageNet, we follow the evaluation setup of ADM \citep{dhariwal2021adm}
and use its reference batches. For text-to-image generation, we compute FID against MS-COCO validation split, and on 50,000 samples for BLIP3-o. The metrics are:
\begin{itemize}[leftmargin=*]
\setlength{\itemsep}{1pt}
\item \textbf{FID} \citep{heusel2017fid} measures the distance between the
feature distributions of real and generated images. It uses the Inception-v3
network \citep{szegedy2016inception} and models both distributions as
Gaussians.
\item \textbf{IS} \citep{salimans2016is} also uses Inception-v3. It is high
when the class prediction for each generated image is confident and the
predictions over all images are diverse.
\item \textbf{Precision and recall} \citep{kynkaanniemi2019pr} measure the
fraction of generated images that fall on the real data manifold, and the
fraction of the real data manifold that the generated images cover,
respectively.
\item \textbf{CLIP score} \citep{hessel2021clipscore} measures how well a
generated image matches its text prompt, as the similarity of the two in the
embedding space of CLIP \citep{radford2021clip}.
\item \textbf{CKNNA and linear probing} measure the semantics of the
diffusion features (Appendix~\ref{app:cknna}).
\end{itemize}

\section{Limitations}
\label{app:limitations}

\textbf{Fidelity--diversity balance.} RECON strongly improves sample
fidelity. At 80 epochs, it increases the
precision of SiT-XL/2 ($0.70 \to 0.81$) and IS ($125.4 \to 204.2$), while
recall decreases ($0.65 \to 0.56$) (Table~\ref{tab:ablation}). Our hunch is
that strong semantic conditioning makes the model commit to clear class
semantics early, which leaves less room for variation. RAE \citep{zheng2025rae}, whose latent space is DINOv2
features, shows the same behavior (precision 0.82 and recall 0.59 at 80
epochs).

\needspace{8\baselineskip}

\section{Discussion and future work}
\label{app:discussion}

\textbf{Combining with self-contained alignment.} Nothing in RECON depends
on where the target representation comes from, only on there being a
trained map into a better-organized space. Pairing it with a method that
derives its target internally \citep{jiang2026sra, haghighi2026layersync,
chefer2026selfflow} would remove the external encoder entirely while
keeping the mechanism intact. We regard this as the most promising
direction.

\textbf{Is the gain bottlenecked by the tokenizer?} RECON improves
generation on all three tokenizers, but by different amounts
(Table~\ref{tab:latents}). FID goes from $7.84 \to 3.25$ on SD-VAE and
$7.41 \to 4.78$ on InVAE, but only from $4.95 \to 4.02$ on VA-VAE. VA-VAE is
the only tokenizer trained with an alignment loss, so its latent already
contains part of the semantics that RECON adds. In other words, RECON helps
most when the latent has no semantic structure of its own. We do not yet know
why the remaining gain on VA-VAE is small. One possibility is that the
tokenizer itself limits how much can be gained. Another is that our semantic
token, a single pooled summary of the image, is not the right signal for a
latent that is already semantically organized, and a signal with information
that the latent does not contain could still help. Telling these two apart
would show whether RECON is close to its limit on such tokenizers, or whether
a different way of injecting semantics could give further gains.

\textbf{Why no time-dependent scheduling during inference?} During inference, for $t < t_{\max}$, the decoder receives the same kind of input as in training. For $t \geq t_{\max}$, we still inject semantic tokens, although $\mathcal{D}_\theta$ does not see the semantic tokens in training, and $g_\phi(h)$ is less accurate at these noise levels.
In practice this improves gFID (Table~\ref{tab:inference-injection}), which suggests that $\mathcal{D}_\theta$ applies what it learned at low noise levels across the whole trajectory.

\needspace{30\baselineskip}
\section{Additional results}
\label{app:extra}

\subsection{Full metrics}
\label{app:fullmetrics}
Fig.~\ref{fig:generalization} reports only FID. Table~\ref{tab:fullmetrics}
gives FID, IS, precision and recall for the same runs, varying the
\hlwrap{cscale}{model scale}, the \hlwrap{ctokz}{tokenizer} and the
\hlwrap{cenc}{target encoder}.

\input{tables/scale_latents}

\needspace{46\baselineskip}
\subsection{Recovering heavily corrupted images}
\label{app:sdedit}

\begin{figure}[!h]
\centering
\includegraphics[width=\linewidth]{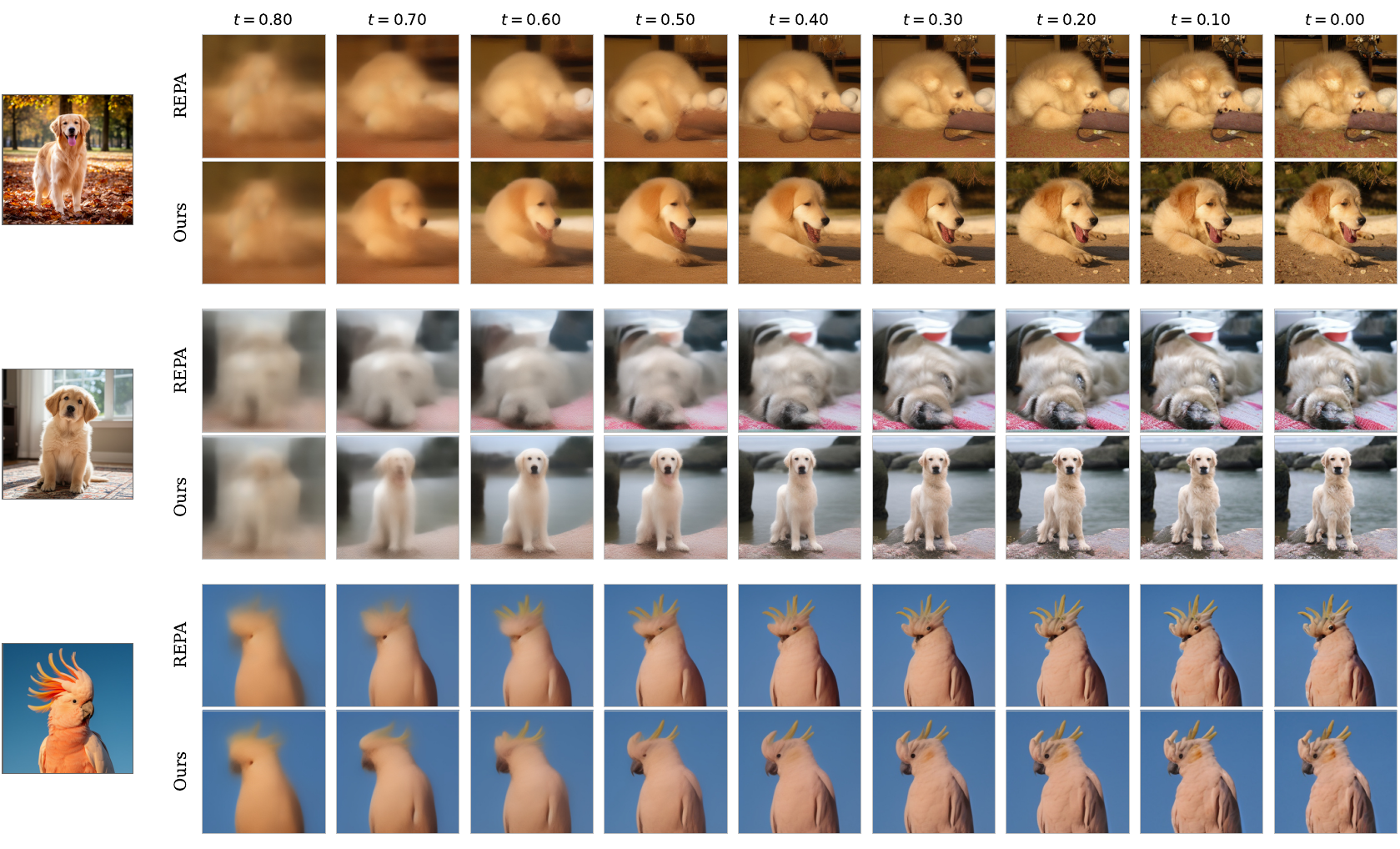}
\caption{\textbf{Denoising heavily corrupted images.} We take three clean
images (left) and corrupt each with heavy noise (80\% noise and 20\% image). Then let REPA and RECON denoise them; each frame
shows the clean-image estimate $\hat{x}_0(t)$. RECON recovers the identity
and layout of the original image better than REPA, which shows its strength in retaining
semantics under heavy noise.}
\label{fig:sdedit}
\end{figure}

In Section~\ref{sec:dynamics}, we show that RECON decides the structure of a
sample earlier in generation. Here we test a related question: can the model
recover an image from heavy corruption? We take three clean images and
corrupt each with heavy noise, moving it along the interpolant to $t=0.8$,
so the input is 80\% noise and 20\% image. Then we let REPA and RECON
(SiT-XL/2) denoise it. Both start from the same noisy latent and share the
class label, the sampling noise and the sampler (Euler--Maruyama, 200 steps,
without CFG), so any difference comes from the model.
Fig.~\ref{fig:sdedit} shows the clean-image estimate $\hat{x}_0(t)$ along
the way. RECON commits to the subject within the first few steps and then
refines it, while the estimates of REPA stay blurry for much longer. As a
result, RECON recovers the identity and layout of the original image better.
Since most of the input is noise, neither model reconstructs it exactly; the
difference is in how fast and how well a plausible image is formed.

\clearpage
\subsection[More qualitative results]{More qualitative
results\footnote{For all the following figures, we use the SiT-XL/2 + RECON
checkpoint at 80 epochs.}}
\label{app:qualitative}

\newsavebox{\gridbox}
\newlength{\gridcell}
\newlength{\gtx}
\newlength{\gty}
\newcommand{\gtile}[2]{%
  \setlength{\gtx}{\dimexpr\gridcell*(1+258*#2)/258\relax}%
  \setlength{\gty}{\dimexpr\gridcell*(1+258*#1)/258\relax}%
  \begin{tikzpicture}
    \useasboundingbox (0,0) rectangle (\gridcell,\gridcell);
    \clip (0,0) rectangle (\gridcell,\gridcell);
    \node[anchor=north west,inner sep=0pt] at (-\gtx,\gridcell+\gty)
      {\usebox{\gridbox}};
  \end{tikzpicture}}
\newcommand{\uncuratedgrid}[3]{%
  \begin{subfigure}{\linewidth}
    \centering
    \setlength{\gridcell}{\dimexpr\linewidth/8\relax}%
    \sbox{\gridbox}{\includegraphics[width=\dimexpr\gridcell*1550/258\relax]%
      {figures/grid_class#1.jpg}}%
    {\offinterlineskip
    \gtile00\gtile01\gtile02\gtile03\gtile04\gtile05\gtile10\gtile11\par
    \gtile12\gtile13\gtile14\gtile15\gtile20\gtile21\gtile22\gtile23\par
    \gtile24\gtile25\gtile30\gtile31\gtile32\gtile33\gtile34\gtile35\par}
    \caption{Class label = ``#2'' (#3)}
  \end{subfigure}}
\newcommand{\uncuratedcaption}{\textbf{Uncurated generation results} of
SiT-XL/2 + RECON. We use classifier-free guidance with $w=4.0$.}

\begin{figure}[!h]
\centering
\uncuratedgrid{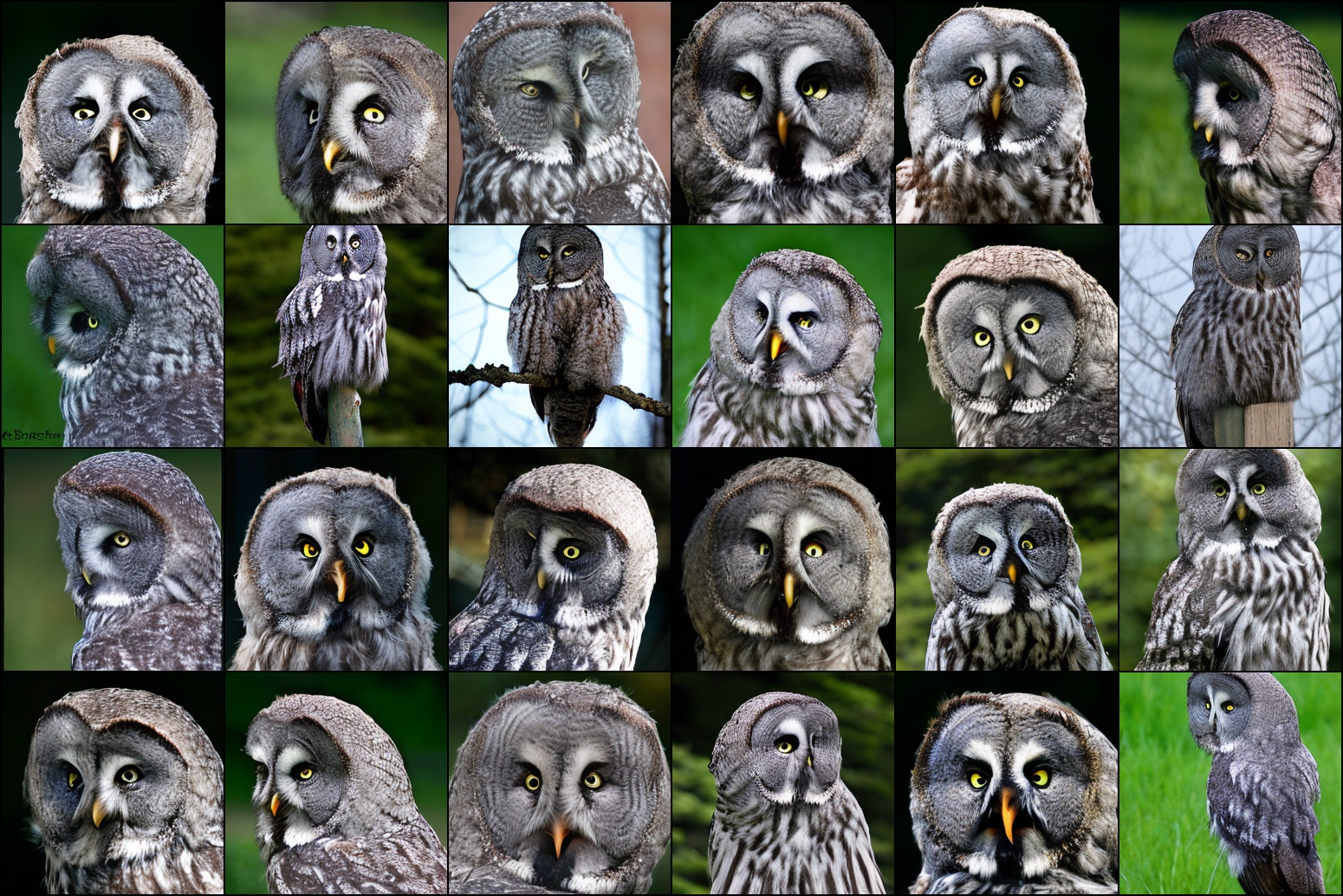}{great grey owl}{24}\par\vspace{6pt}
\uncuratedgrid{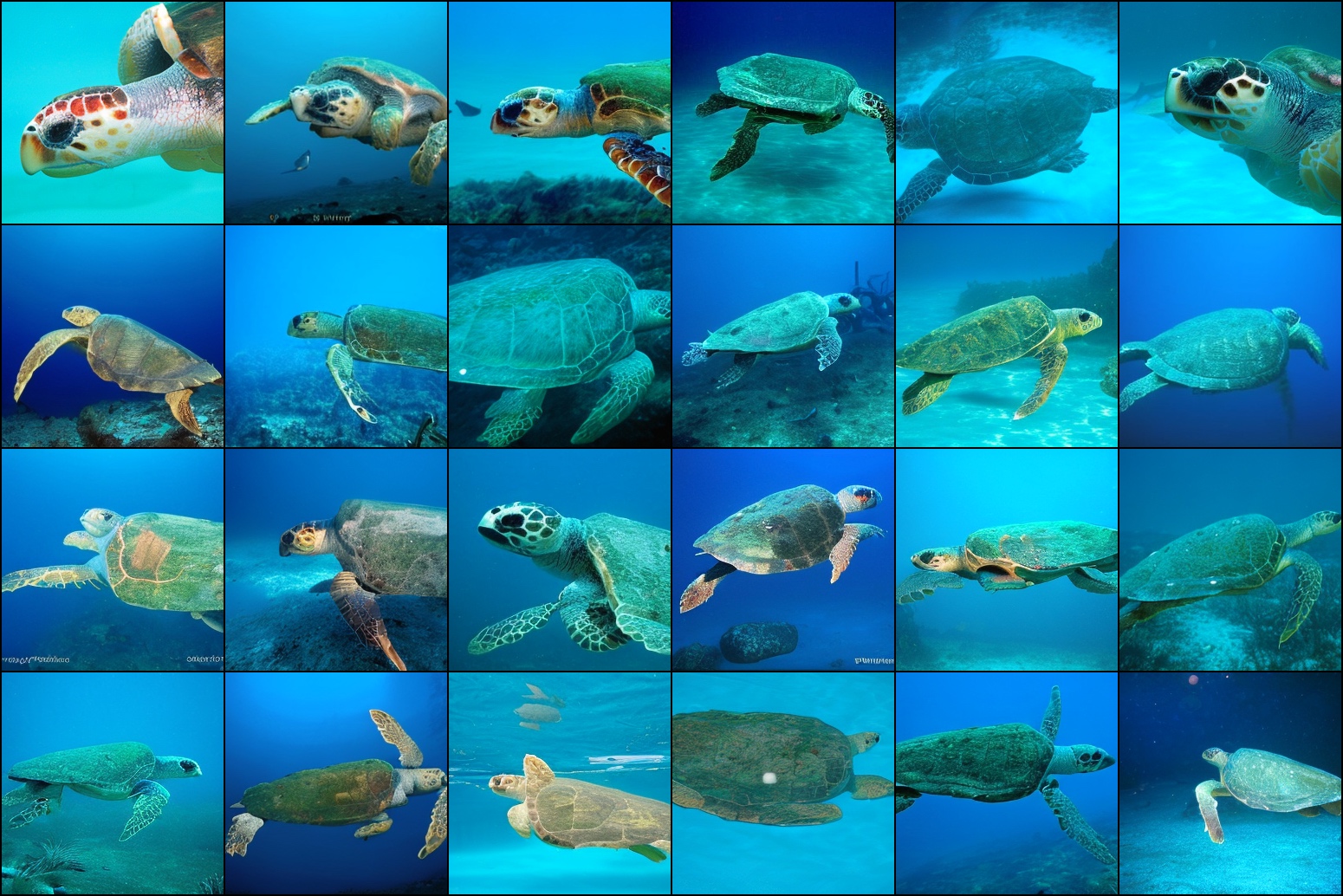}{loggerhead turtle}{33}\par\vspace{6pt}
\uncuratedgrid{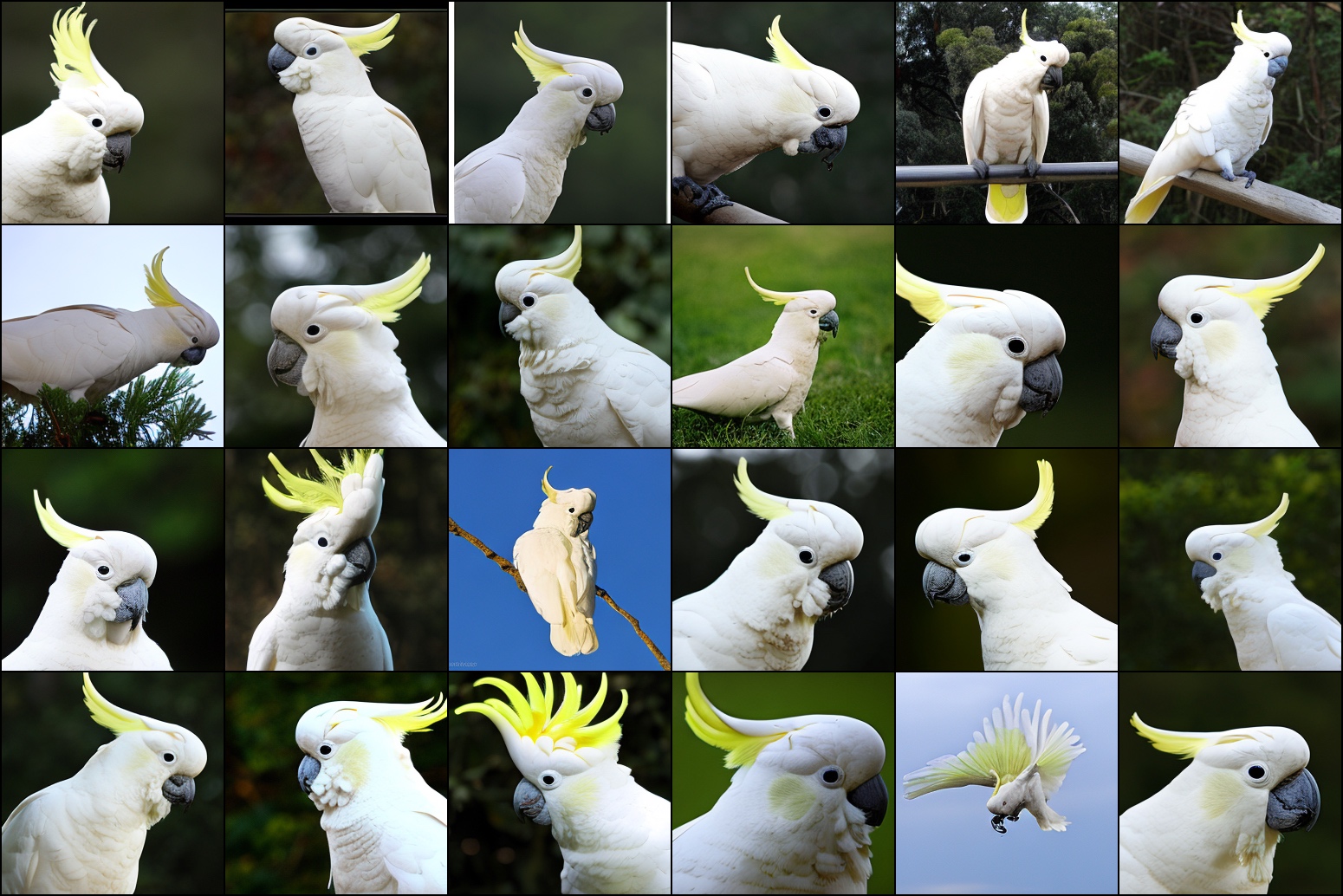}{sulphur-crested cockatoo}{89}
\caption{\uncuratedcaption}
\label{fig:uncurated1}
\end{figure}

\begin{figure}[p]
\centering
\uncuratedgrid{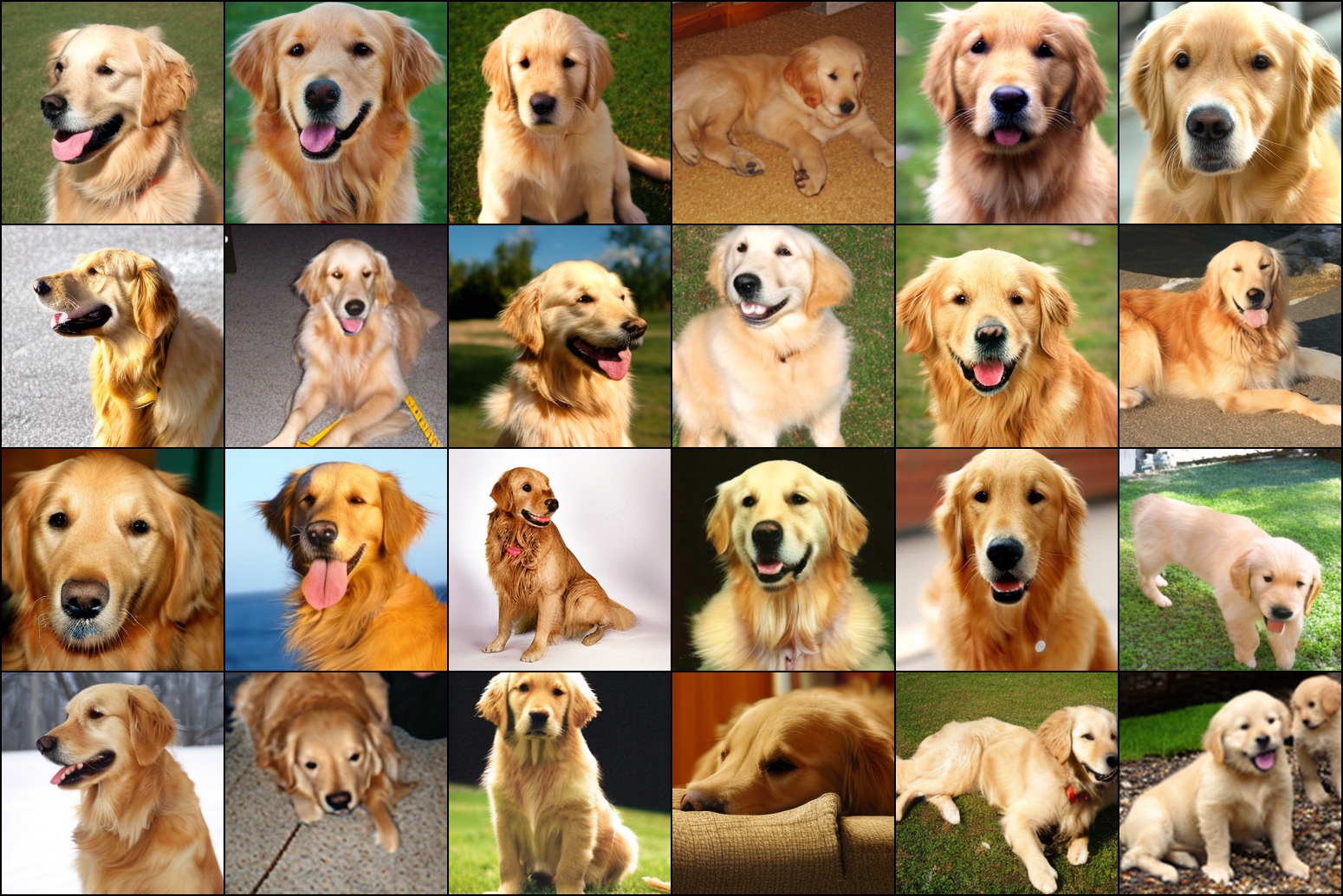}{golden retriever}{207}\par\vspace{6pt}
\uncuratedgrid{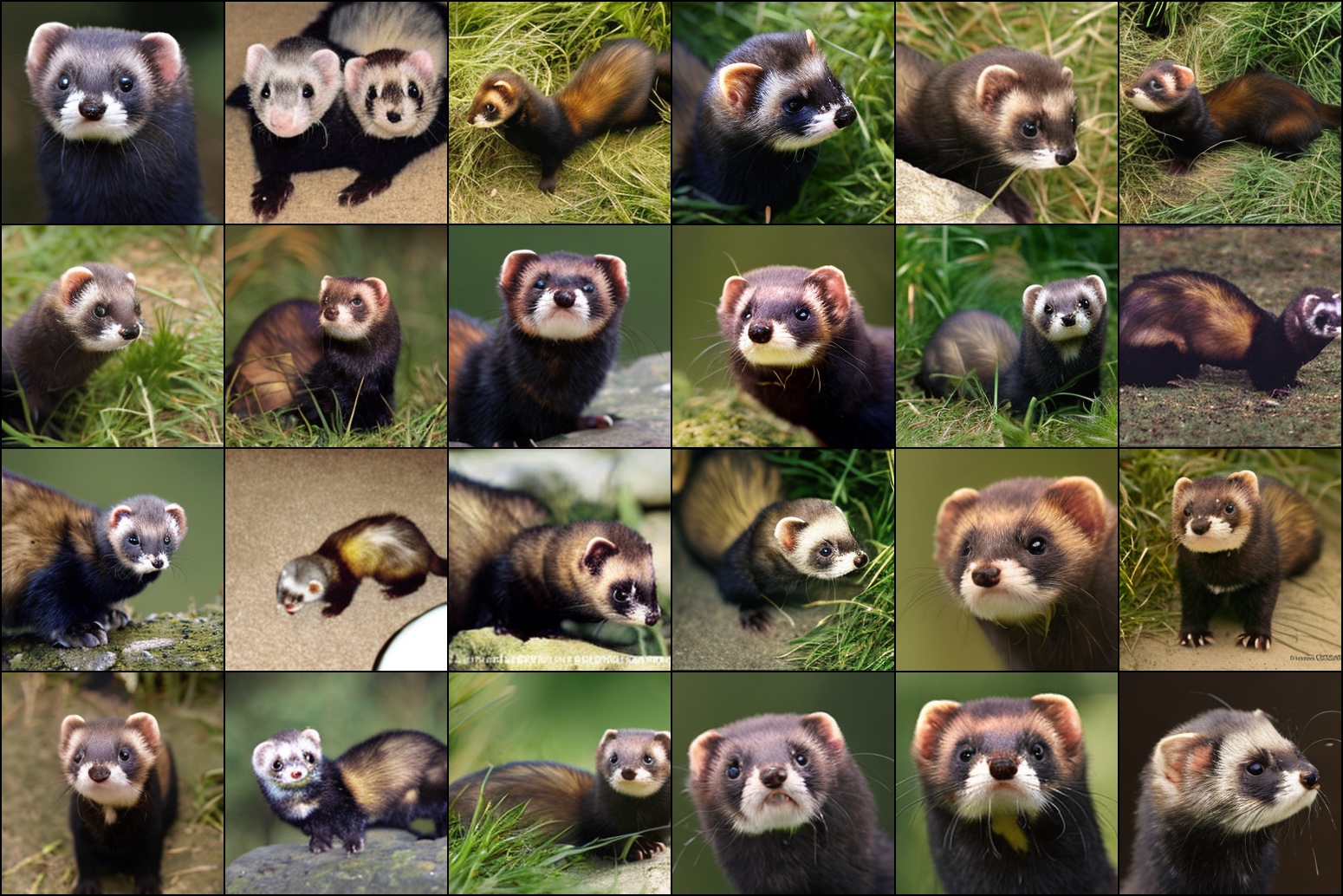}{polecat}{358}\par\vspace{6pt}
\uncuratedgrid{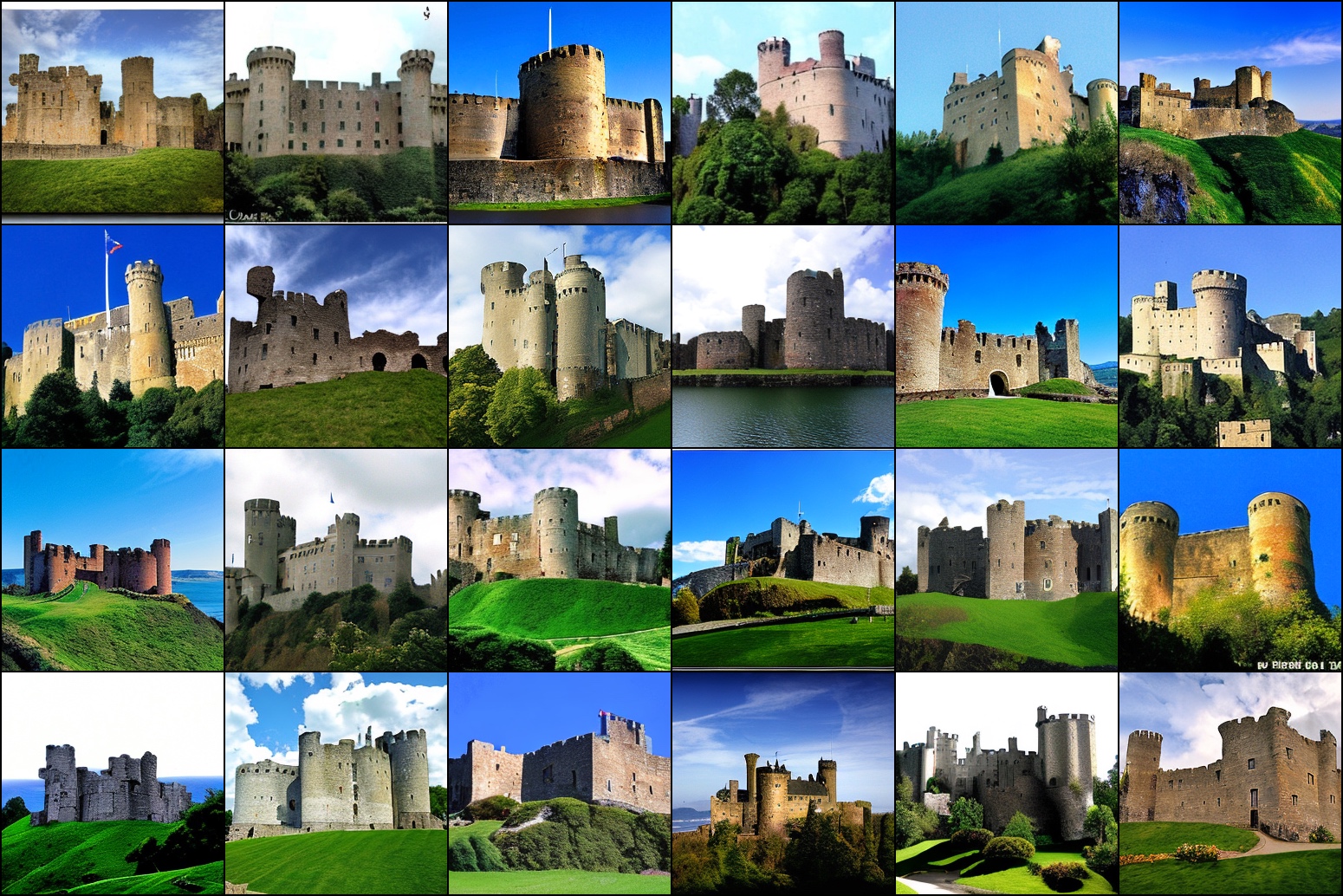}{castle}{483}
\caption{\uncuratedcaption}
\label{fig:uncurated2}
\end{figure}

\begin{figure}[p]
\centering
\uncuratedgrid{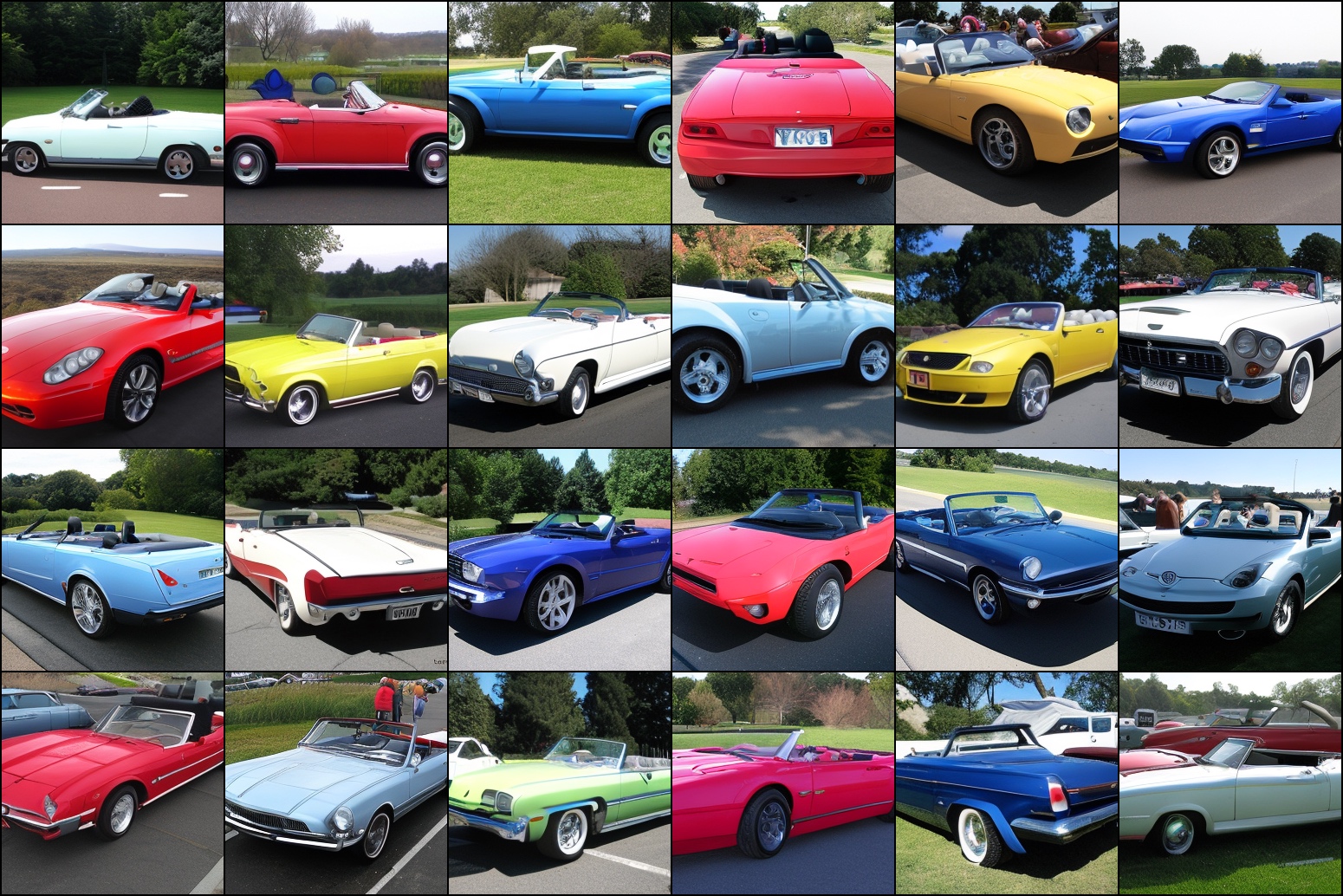}{convertible}{511}\par\vspace{6pt}
\uncuratedgrid{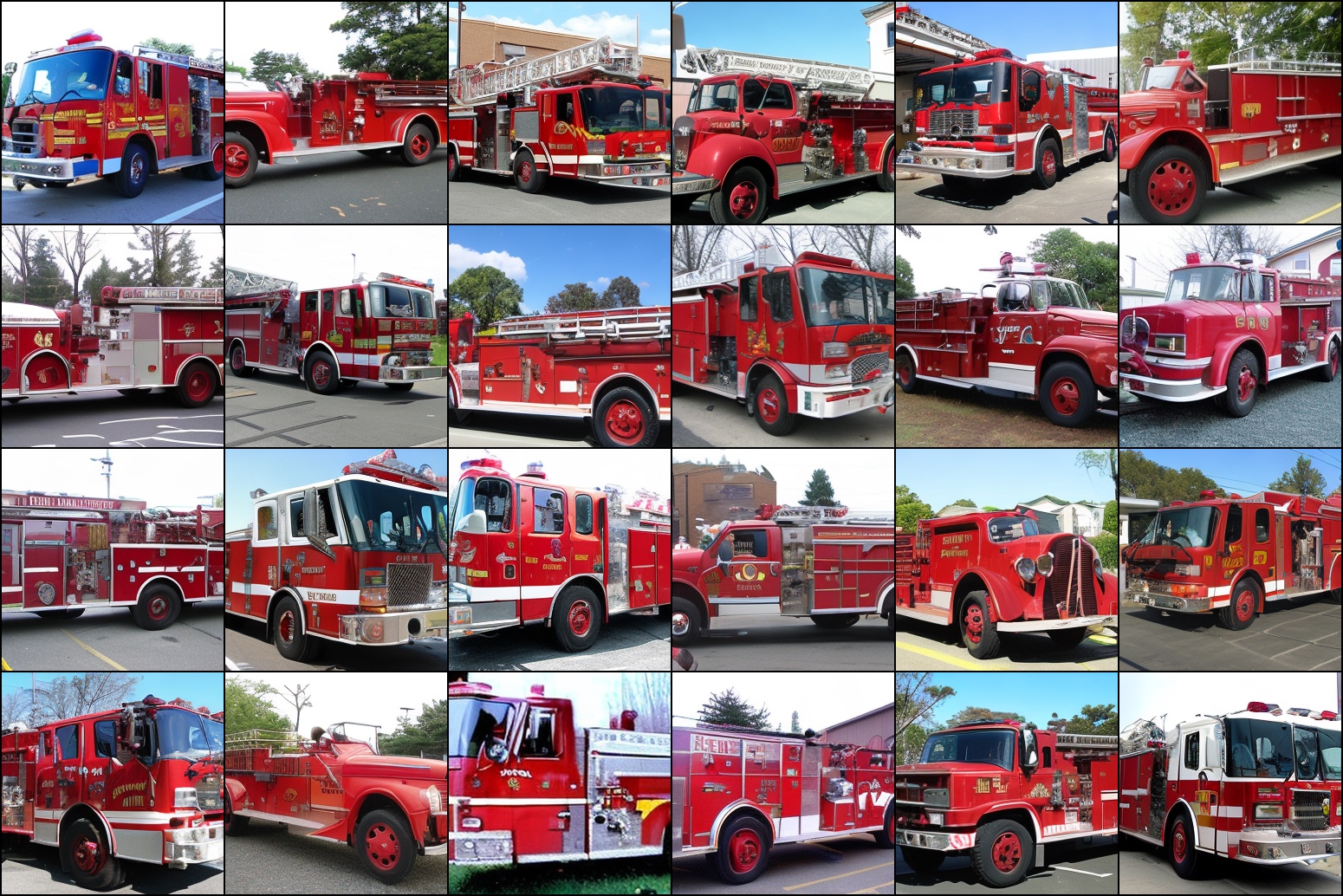}{fire engine}{555}\par\vspace{6pt}
\uncuratedgrid{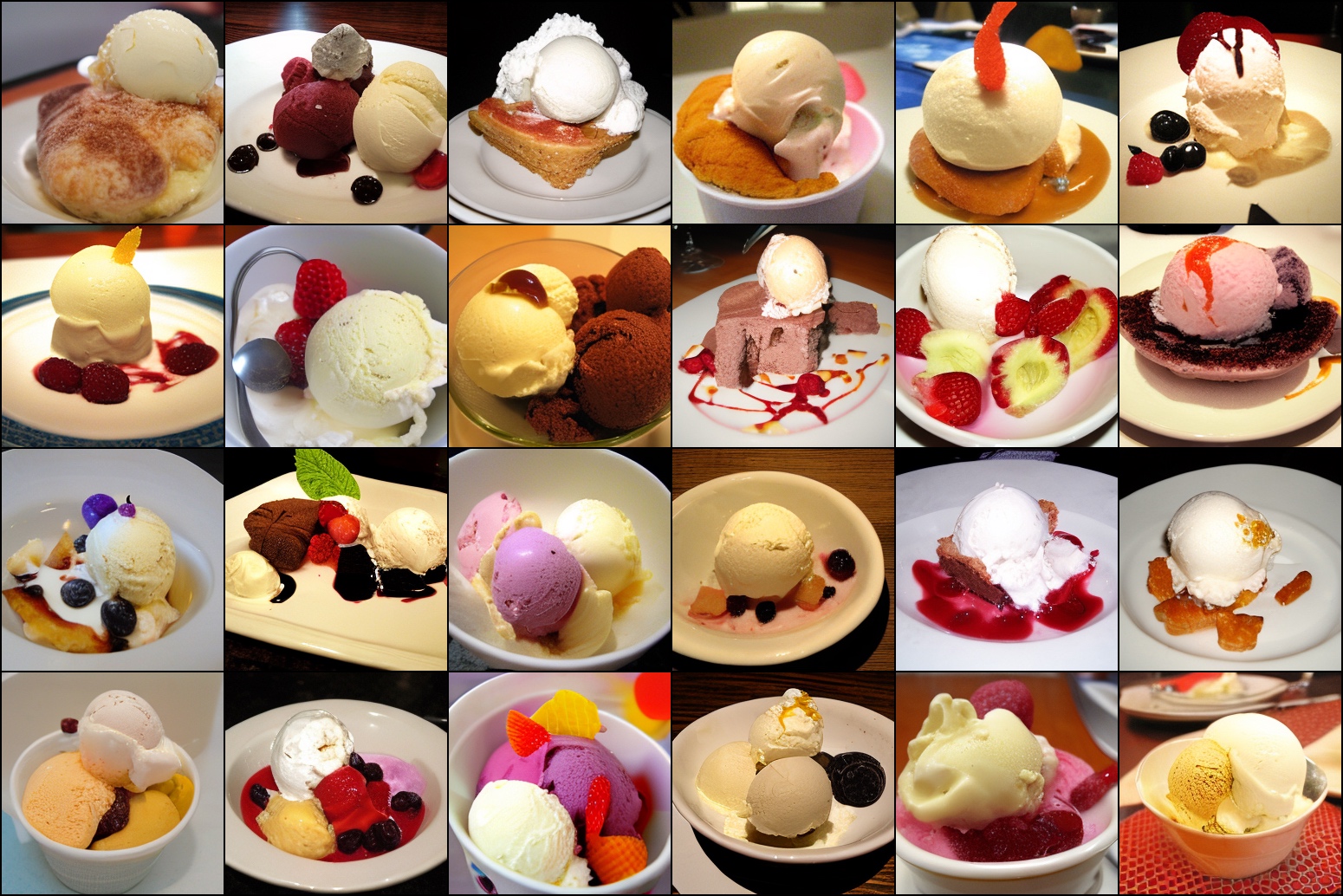}{ice cream}{928}
\caption{\uncuratedcaption}
\label{fig:uncurated3}
\end{figure}

\begin{figure}[p]
\centering
\uncuratedgrid{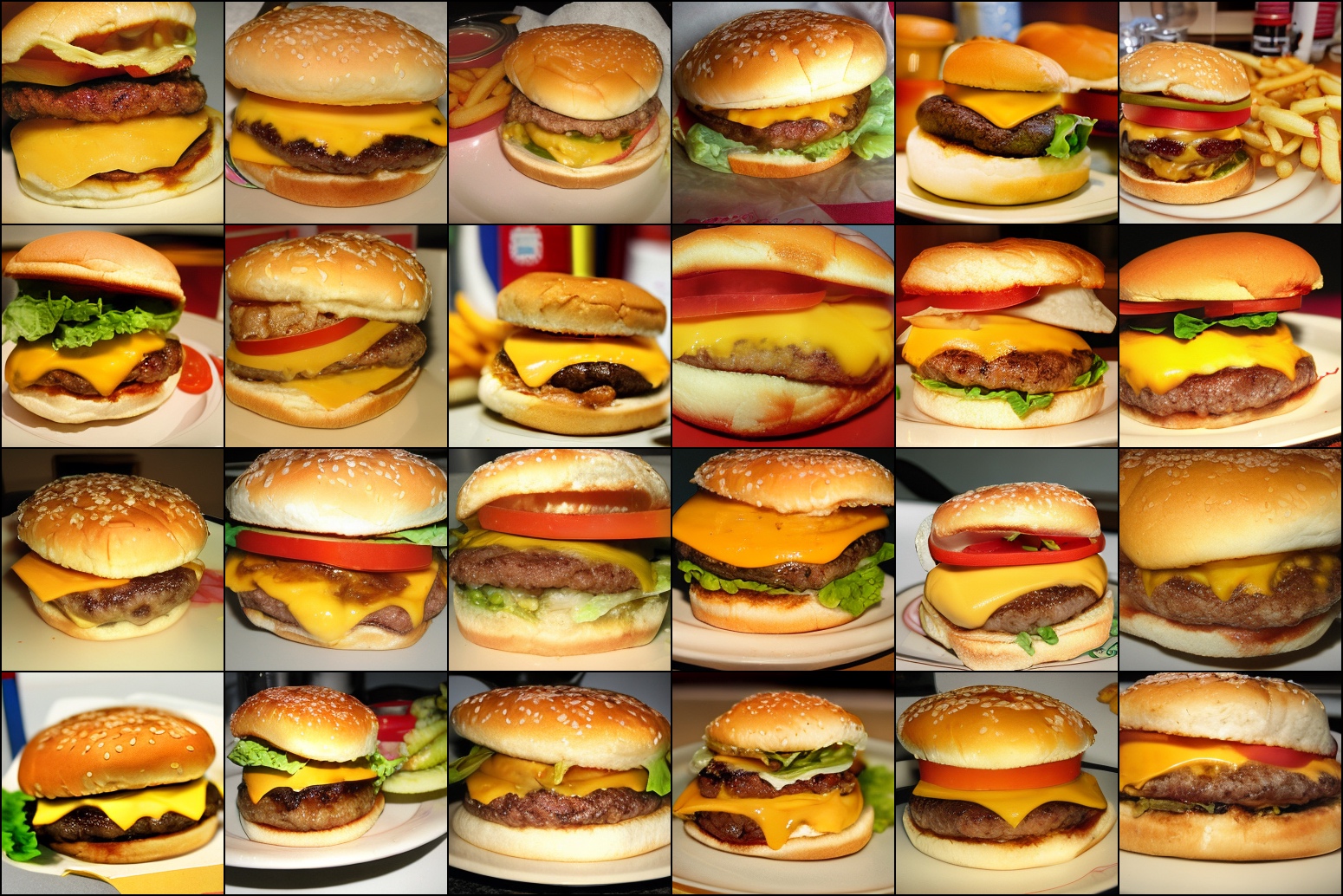}{cheeseburger}{933}\par\vspace{6pt}
\uncuratedgrid{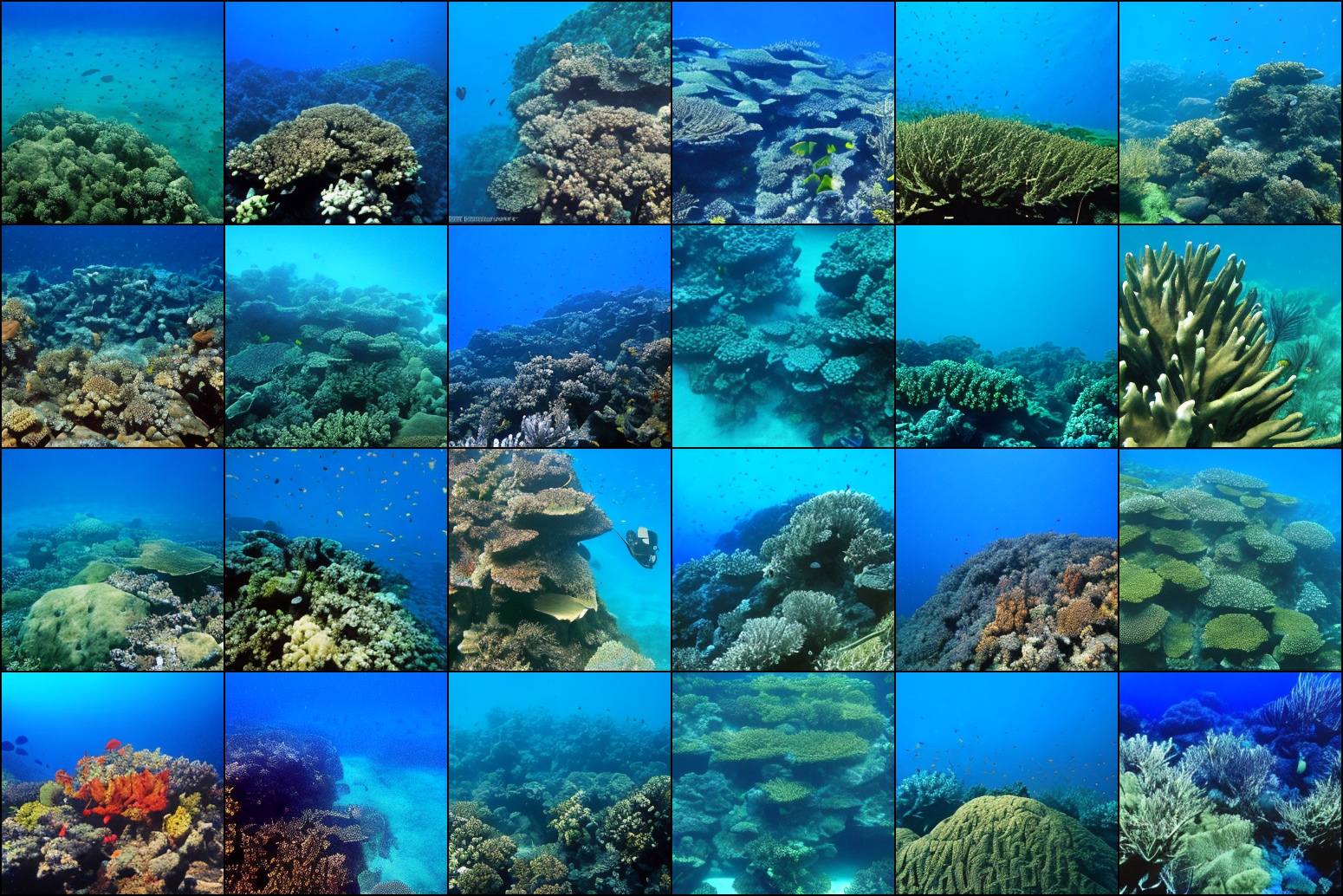}{coral reef}{973}\par\vspace{6pt}
\uncuratedgrid{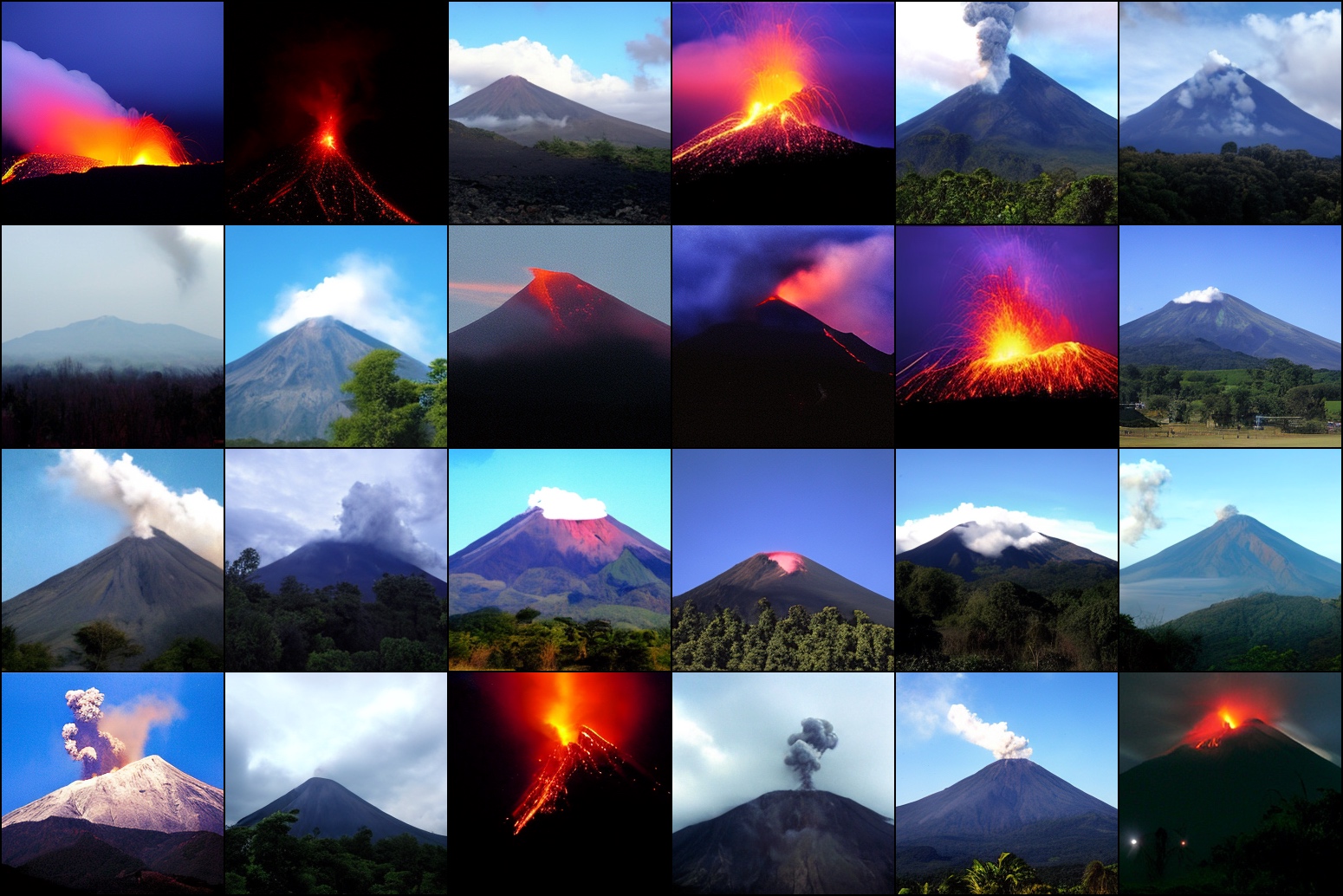}{volcano}{980}
\caption{\uncuratedcaption}
\label{fig:uncurated4}
\end{figure}

\end{document}

%% file: tables/nocfg_system.tex
\begin{table}[t]
    \begin{minipage}[t]{0.40\textwidth}
    \caption{\textbf{gFID Comparison without CFG} on ImageNet $256\times256$. SD-VAE latents by default. $\dagger$: VA-VAE latents.}
    \label{tab:nocfg}
    \end{minipage}\hfill
    \begin{minipage}[t]{0.58\textwidth}
    \caption{\textbf{System-level comparison} on ImageNet $256\times256$ with
    guidance. RECON: Autoguidance~\citep{karras2024autoguidance}
    ($\ddagger$: improved, Appendix~\ref{app:guidance}); baselines: CFG
    (*: scheduled).}
    \label{tab:system}
    \end{minipage}\par\vspace{-1pt}
    \begin{minipage}[t]{0.40\textwidth}
    \centering
    \scriptsize
    \setlength{\tabcolsep}{5.26pt}
    \begin{tabular}{lcc}
    \toprule
    Method & Iter. & gFID$\downarrow$ \\
    \midrule
    \multicolumn{3}{l}{\emph{No representation alignment}} \\
    DiT-XL/2   & 7M   & 9.60 \\
    SiT-XL/2   & 7M   & 8.30 \\
    MDT~\citep{gao2023mdt}        & 6.5M & 6.23 \\
    MaskDiT~\citep{zheng2024maskdit}    & 8M   & 5.69 \\
    \midrule
    \multicolumn{3}{l}{\emph{Alignment w/o external encoder}} \\
    SRA        & 4M & 7.27 \\
    LayerSync  & 4M & 6.32   \\
    Self-Flow  & 4M & 5.70 \\
    \midrule
    \multicolumn{3}{l}{\emph{Alignment w/ external encoder}} \\
    REPA         & 400K & 7.90 \\
    REPA         & 4M   & 5.90 \\
    REG          & 400K & 3.40 \\
    REG          & 1M & 2.70 \\
    LightningDiT$^\dagger$~\citep{yao2025vavae} & 400K & 4.29 \\
    REPA-E       & 400K & 4.07 \\
    \rowcolor{gray!12}
    RECON        & 100K & 9.25   \\
    \rowcolor{gray!12}
    RECON        & 200K & 4.64   \\
    \rowcolor{gray!12}
    RECON        & 400K & 3.25   \\
    \rowcolor{gray!12}
    RECON        & 1M   & 2.54   \\
    \rowcolor{gray!12}
    RECON        & 2.4M   & 2.21   \\
    \bottomrule
    \end{tabular}
    \end{minipage}\hfill
    \begin{minipage}[t]{0.58\textwidth}
    \centering
    \scriptsize
    \setlength{\tabcolsep}{1.90pt}
    \begin{tabular}{lccccc}
    \toprule
    Model & Epochs & FID$\downarrow$ & IS$\uparrow$ & Pre.$\uparrow$ & Rec.$\uparrow$ \\
    \midrule
    \multicolumn{6}{l}{\emph{Pixel diffusion}} \\
    ADM-G + ADM-U~\citep{dhariwal2021adm}            & 400  & 3.94 & 186.7 & 0.82 & 0.52 \\
    VDM++~\citep{kingma2023vdmpp}            & 560  & 2.40  & 225.3 & --   & --   \\
    Simple diffusion~\citep{hoogeboom2023simple} & 800  & 2.77  & 211.8 & --   & --   \\
    CDM~\citep{ho2022cdm}              & 2160 & 4.88 & 158.7 & --   & --   \\
    \midrule
    \multicolumn{6}{l}{\emph{Latent diffusion, U-Net}} \\
    LDM-4~\citep{rombach2022ldm}            & 200  & 3.60  & 247.7 & 0.87 & 0.48 \\
    \midrule
    \multicolumn{6}{l}{\emph{Latent diffusion, Transformer + U-Net hybrid}} \\
    U-ViT-H/2~\citep{bao2023uvit}        & 240  & 2.29 & 263.9 & 0.82 & 0.57 \\
    DiffiT*~\citep{hatamizadeh2024diffit}          & --   & 1.73  & 276.5 & 0.80 & 0.62 \\
    MDTv2-XL/2*      & 1080 & 1.58 & 314.7 & 0.79 & 0.65 \\
    \midrule
    \multicolumn{6}{l}{\emph{Latent diffusion, Transformer}} \\
    MaskDiT          & 1600 & 2.28 & 276.6 & 0.80 & 0.61 \\
    SD-DiT~\citep{zhu2024sddit}           & 480  & 3.23  & --    & --   & --   \\
    DiT-XL/2         & 1400 & 2.27 & 278.2 & 0.83 & 0.57 \\
    SiT-XL/2         & 1400 & 2.06 & 270.3 & 0.82 & 0.59 \\
    \quad + REPA     & 800  & 1.80 & 284.0 & 0.81 & 0.61 \\
    \quad + REPA*    & 800  & 1.42 & 305.7 & 0.80 & 0.65 \\
    \quad + REG      & 480  & 1.40 & 296.9 & 0.77 & 0.66 \\
    \rowcolor{gray!12}
    \quad + RECON  & 80   & 2.26   & 241.22  & 0.82 & 0.57 \\
    \rowcolor{gray!12}
    \quad + RECON$^\ddagger$ & 80 & 2.09 & 224.20 & 0.79 & 0.60 \\
    \rowcolor{gray!12}
    \quad + RECON  & 480   & 1.80   & 273.30  & 0.81  & 0.59   \\
    \rowcolor{gray!12}
    \quad + RECON$^\ddagger$ & 480 & 1.49 & 269.10 & 0.79 & 0.62 \\
    \bottomrule
    \end{tabular}
    \end{minipage}
    \end{table}

%% file: tables/t2i_row.tex
\begin{table}[!ht]
\centering
\begin{minipage}[t]{0.505\textwidth}
\caption{\textbf{Text-to-image on MS-COCO.} MMDiT, ODE, 50 NFE, 150K iter.;
$w$: guidance scale. RECON is superior to REPA.}
\label{tab:coco}
\end{minipage}\hfill
\begin{minipage}[t]{0.470\textwidth}
\caption{\textbf{Text-to-image on BLIP3-o} \citep{chen2025blip3o}\textbf{.}
NanoGen \citep{leng2026diffusionbench}; ODE, 50 NFE, 100K iter.}
\label{tab:blip3o}
\end{minipage}\par\vspace{-6pt}
\footnotesize
\setlength{\tabcolsep}{3pt}
\setlength{\aboverulesep}{0.3ex}\setlength{\belowrulesep}{0.35ex}
\begin{minipage}[c]{0.505\textwidth}
\centering
\setlength{\tabcolsep}{3.20pt}%
\begin{tabular}{lcccc}
\toprule
& \multicolumn{2}{c}{$w{=}1$} & \multicolumn{2}{c}{$w{=}2$} \\
\cmidrule(lr){2-3}\cmidrule(l){4-5}
Method & FID$\downarrow$ & CLIP$\uparrow$ & FID$\downarrow$ & CLIP$\uparrow$ \\
\midrule
+ REPA & 9.85 & 0.2273 & 4.73 & 0.2443 \\
\rowcolor{gray!12}
+ RECON (ours) & 8.26 & 0.2313 & 4.55 & 0.2462 \\
\bottomrule
\end{tabular}
\end{minipage}\hfill
\begin{minipage}[c]{0.470\textwidth}
\centering
\setlength{\tabcolsep}{4.65pt}%
\begin{tabular}{lcccc}
\toprule
& \multicolumn{2}{c}{$w{=}1$} & \multicolumn{2}{c}{$w{=}8$} \\
\cmidrule(lr){2-3}\cmidrule(l){4-5}
Method & FID$\downarrow$ & IS$\uparrow$ & FID$\downarrow$ & IS$\uparrow$ \\
\midrule
+ REPA & 5.41 & 22.40 & 5.30 & 27.61 \\
\rowcolor{gray!12}
+ RECON (ours) & 3.90 & 24.31 & 5.04 & 27.91 \\
\bottomrule
\end{tabular}
\end{minipage}
\end{table}

%% file: tables/ablation_design.tex
\newsavebox{\ablationbox}
\begin{table}[!h]
\captionsetup{font=small}
\savebox{\ablationbox}{%
\begin{minipage}[t]{0.675\linewidth}
\caption{\textbf{Component-wise analysis} on ImageNet $256\times256$. All
models are SiT-XL/2 with DINOv2-B as the target, trained for 400K iterations
and measured without classifier-free guidance. Each block varies a single
component, shaded, with all others held at the default.}
\label{tab:ablation}
\centering
\footnotesize
\setlength{\tabcolsep}{3pt}
\begin{tabular*}{\linewidth}{@{\extracolsep{\fill}}ccccrrcc@{}}
\toprule
Depth & $k$ & $t_{\max}$ & DM mech. & FID$\downarrow$ & IS$\uparrow$ & Pre.$\uparrow$ & Rec.$\uparrow$ \\
\midrule
\multicolumn{4}{@{}l}{Vanilla SiT-XL/2 \citep{ma2024sit}} & 17.20 & 76.52 & 0.65 & 0.63 \\
\multicolumn{4}{@{}l}{SiT-XL/2 + REPA \citep{yu2025repa}} & 7.83 & 125.43 & 0.70 & 0.65 \\
\midrule
\vary[cdepth]{6}  & 1 & 0.5 & zero-pad & 3.26 & 205.86 & 0.81 & 0.56 \\
\vary[cdepth]{8}  & 1 & 0.5 & zero-pad & 3.25 & 204.18 & 0.81 & 0.56 \\
\vary[cdepth]{10} & 1 & 0.5 & zero-pad & 3.38 & 202.34 & 0.81 & 0.55 \\
\midrule
8 & \vary[ctok]{1}  & 0.5 & zero-pad & 3.25 & 204.18 & 0.81 & 0.56 \\
8 & \vary[ctok]{4}  & 0.5 & zero-pad & 4.70 & 214.01 & 0.84 & 0.50 \\
\midrule
8 & 1 & \vary[ctmax]{0.3} & zero-pad & 5.33 & 146.74 & 0.74 & 0.63 \\
8 & 1 & \vary[ctmax]{0.5} & zero-pad & 3.25 & 204.18 & 0.81 & 0.56 \\
8 & 1 & \vary[ctmax]{1.0} & zero-pad & 4.70 & 236.82 & 0.84 & 0.44 \\
\midrule
8 & 1 & 0.5 & \vary[cdm]{zero-pad} & 3.25 & 204.18 & 0.81 & 0.55 \\
8 & 1 & 0.5 & \vary[cdm]{linear}   & 3.68 & 198.35 & 0.80 & 0.56 \\
\bottomrule
\end{tabular*}
\end{minipage}}%
\usebox{\ablationbox}\hfill
\begin{minipage}[t][\dimexpr\ht\ablationbox+\dp\ablationbox\relax][s]{0.300\linewidth}
\caption{\textbf{Teacher forcing vs.\ other training strategies.}
`online'/`EMA': tokens from $g_\phi$ or its moving average. gFID and IS
after 100K iterations.}
\label{tab:tokens}
\centering
\footnotesize
\setlength{\tabcolsep}{3pt}
\begin{tabular}{@{}lcc@{}}
\toprule
Tokens built from & gFID$\downarrow$ & IS$\uparrow$ \\
\midrule
\rowcolor{gray!12}
$f_{\text{ext}}(x_0)$                  & 9.2 & 118.8 \\
$g_\phi(h)$, online                    & 17.1 & 76.3 \\
$g_\phi(h)$, EMA                       & 16.8 & 77.5 \\
\bottomrule
\end{tabular}
\vfill
\caption{\textbf{Injection at inference.} Semantic tokens only for
$t<t_{\max}$, as in training, or at every sampling step.}
\label{tab:inference-injection}
\centering
\begin{tabular}{@{}lcc@{}}
\toprule
Injected at & gFID$\downarrow$ & IS$\uparrow$ \\
\midrule
$t < t_{\max}$ only & 6.3 & 149.9 \\
\rowcolor{gray!12}
All timesteps     & 3.2 & 202.2 \\
\bottomrule
\end{tabular}\par\vspace{0pt}%
\end{minipage}
\end{table}

%% file: tables/scale_latents.tex
\begin{table}[!h]
\centering
\footnotesize
\setlength{\tabcolsep}{4.5pt}
\newcommand{\cs}{\cellcolor{cscale}}
\newcommand{\ct}{\cellcolor{ctokz}}
\newcommand{\ce}{\cellcolor{cenc}}
\caption{\textbf{Full metrics} for Fig.~\ref{fig:generalization}. ImageNet
$256\times256$, no CFG.}
\label{tab:fullmetrics}\label{tab:scale}\label{tab:latents}\label{tab:encoders}
\begin{tabular}{lcccccccc}
\toprule
 & \multicolumn{4}{c}{REPA} & \multicolumn{4}{c}{RECON} \\
\cmidrule(lr){2-5}\cmidrule(lr){6-9}
 & FID$\downarrow$ & IS$\uparrow$ & Pre.$\uparrow$ & Rec.$\uparrow$
 & FID$\downarrow$ & IS$\uparrow$ & Pre.$\uparrow$ & Rec.$\uparrow$ \\
\midrule
\multicolumn{9}{l}{\textit{Model scale}: 20 epochs, SD-VAE latents} \\
SiT-B/2  & 46.35 & 30.53 & 0.477 & \textbf{0.608} & \cs\textbf{26.08} & \cs\textbf{54.63} & \cs\textbf{0.620} & \cs 0.577 \\
SiT-L/2  & 22.95 & 59.17 & 0.618 & \textbf{0.610} & \cs\textbf{11.24} & \cs\textbf{104.44} & \cs\textbf{0.742} & \cs 0.550 \\
SiT-XL/2 & 18.88 & 69.34 & 0.647 & \textbf{0.608} & \cs\textbf{9.25} & \cs\textbf{118.80} & \cs\textbf{0.760} & \cs 0.543 \\
\midrule
\multicolumn{9}{l}{\textit{Tokenizer}: SiT-XL/2, 80 epochs} \\
SD-VAE & 7.84 & 125.43 & 0.700 & \textbf{0.654} & \ct\textbf{3.25} & \ct\textbf{204.18} & \ct\textbf{0.817} & \ct 0.565 \\
VA-VAE & 4.95 & 146.43 & 0.761 & \textbf{0.611} & \ct\textbf{4.02} & \ct\textbf{163.04} & \ct\textbf{0.774} & \ct 0.604 \\
InVAE  & 7.41 & 125.25 & 0.712 & \textbf{0.655} & \ct\textbf{4.78} & \ct\textbf{151.10} & \ct\textbf{0.755} & \ct 0.627 \\
\midrule
\multicolumn{9}{l}{\textit{Target encoder}: SiT-B/2, 20 epochs, SD-VAE latents} \\
DINOv2-B       & 46.35 & 30.53 & 0.477 & \textbf{0.608} & \ce\textbf{26.08} & \ce\textbf{54.63} & \ce\textbf{0.620} & \ce 0.577 \\
I-JEPA-H       & 48.56 & 28.41 & 0.465 & \textbf{0.596} & \ce\textbf{26.56} & \ce\textbf{54.36} & \ce\textbf{0.592} & \ce 0.574 \\
MoCo-v3-B      & 51.98 & 26.60 & 0.449 & \textbf{0.597} & \ce\textbf{34.45} & \ce\textbf{41.88} & \ce\textbf{0.584} & \ce 0.547 \\
SigLIP2-B      & 53.85 & 25.12 & 0.440 & \textbf{0.587} & \ce\textbf{43.65} & \ce\textbf{31.29} & \ce\textbf{0.508} & \ce 0.550 \\
WebSSL-DINO-1B & 49.61 & 28.52 & 0.459 & \textbf{0.596} & \ce\textbf{24.92} & \ce\textbf{55.74} & \ce\textbf{0.656} & \ce 0.527 \\
\bottomrule
\end{tabular}
\end{table}